\documentclass[iicol]{sn-jnl2}

\usepackage[ruled,linesnumbered,noend,algo2e]{algorithm2e}
\usepackage{comment}
\usepackage{amsmath} 
\usepackage{subcaption}
\usepackage{bbding}
\usepackage{booktabs}
\usepackage{multirow}
\usepackage{colortbl}
\usepackage{url}
\usepackage{natbib}
\setcitestyle{authoryear,open={(},close={)}} 
\usepackage[figuresright]{rotating}
\usepackage{enumerate}
\usepackage{framed}

\jyear{2021}%

\theoremstyle{thmstyleone}%
\theoremstyle{thmstyletwo}%

\theoremstyle{thmstylethree}%

\begin{document}

\title[Article Title]{Delving into Identify-Emphasize Paradigm for\\ Combating Unknown Bias}


\author[1]{\fnm{Bowen} \sur{Zhao}}\email{zbw18@mails.tsinghua.edu.cn}

\author[2]{\fnm{Chen} \sur{Chen}}\email{chen1634chen@gmail.com}

\author[1,3]{\fnm{Qian-Wei} \sur{Wang}}\email{wanggw21@mails.tsinghua.edu.cn}

\author[2]{\fnm{Anfeng} \sur{He}}\email{alexfenghe@tencent.com}

\author[1,3]{\fnm{Shu-Tao} \sur{Xia}}\email{xiast@sz.tsinghua.edu.cn}

\affil[1]{\orgdiv{Tsinghua Shenzhen International Graduate School}, \orgname{Tsinghua University}, \orgaddress{\country{China}}}

\affil[2]{\orgdiv{TEG AI}, \orgname{Tencent}, \orgaddress{ \country{China}}}

\affil[3]{\orgdiv{Research Center of Artificial Intelligence}, \orgname{Peng Cheng Laboratory}, \orgaddress{ \country{China}}}


\abstract{
Dataset biases are notoriously detrimental to model robustness and generalization.
The identify-emphasize paradigm appears to be effective in dealing with unknown biases.
However, we discover that it is still plagued by two challenges: A, the quality of the identified bias-conflicting samples is far from satisfactory; B, the emphasizing strategies only produce suboptimal performance.
In this paper, for challenge A, we propose an effective bias-conflicting scoring method (ECS) to boost the identification accuracy, along with two practical strategies --- peer-picking and epoch-ensemble.
For challenge B, we point out that the gradient contribution statistics can be a reliable indicator to inspect whether the optimization is dominated by bias-aligned samples. Then, we propose gradient alignment (GA), which employs gradient statistics to balance the contributions of the mined bias-aligned and bias-conflicting samples dynamically throughout the learning process, forcing models to leverage intrinsic features to make fair decisions.
Furthermore, we incorporate self-supervised (SS) pretext tasks into training, which enable models to exploit richer features rather than the simple shortcuts, resulting in more robust models.
Experiments are conducted on multiple datasets in various settings, demonstrating that the proposed solution can mitigate the impact of unknown biases and achieve state-of-the-art performance.
}




\maketitle

\section{Introduction}
Deep Neural Networks (DNNs) have made significant advances in a variety of visual tasks. DNNs tend to learn \textbf{intended} decision rules to accomplish target tasks commonly. However, they may follow \textbf{unintended} decision rules based on the easy-to-learn shortcuts to ``achieve" target goals in some scenarios~\citep{bahng2020learning}. 
For instance, when training a model to classify digits on Colored MNIST~\citep{kim2019learning}, where the images of each class are primarily dyed by one pre-defined color respectively (\textit{e.g.}, most `0' are red, `1' are yellow, see examples in Figure~\ref{fig:examples}), the intended decision rules classify images based on the shape of digits, whereas the unintended decision rules utilize color information instead.
Following~\cite{nam2020learning}, sample $x$ that can be ``correctly" classified by unintended decision rules is denoted as a \textbf{bias-aligned} sample $\underline{x}$ (\textit{e.g.}, red `0' in Colored MNIST) and vice versa a \textbf{bias-conflicting} sample $\overline{x}$ (\textit{e.g.}, green `0').

\begin{figure*}[t]
\centering
\includegraphics[width=0.95\textwidth]{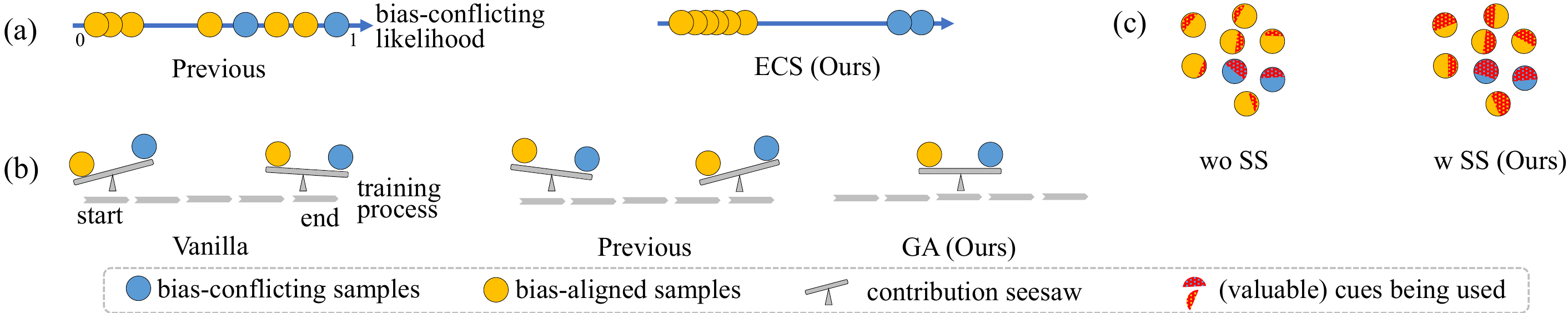}
\caption{(a) Effective bias-Conflicting Scoring (ECS) helps identify real bias-conflicting samples in stage \uppercase\expandafter{\romannumeral1}. (b) Gradient Alignment (GA) balances contributions from the mined bias-aligned and bias-conflicting samples throughout training, enforcing models to focus on intrinsic features in stage \uppercase\expandafter{\romannumeral2}. (c) Self-Supervised (SS) pretext tasks further assist models in capturing more general and robust representations from diverse valuable cues in stage \uppercase\expandafter{\romannumeral2}.
}
\label{fig:intro_effect} 
\end{figure*}

There are many similar scenarios in the real world. For example, an animal-centric image set may be biased by the habitats in the background, and a human-centric set may be biased by gender or racial information. Models blinded by biased datasets usually perform poorly in mismatched distributions (\textit{e.g.}, a red `8' may be incorrectly classified as `0' by the model trained on Colored MNIST). Worse, models with racial or gender bias, \textit{etc.} can cause severe negative social impacts. 
Furthermore, in most real-world problems, the bias information (both bias type and precise labels of bias attribute) is unknown, making debiasing more challenging. Therefore, combating unknown biases is urgently demanded when deploying AI systems in realistic applications.

One major issue that leads to biased models is that the training objective (\textit{e.g.}, vanilla empirical risk minimization) can be accomplished through only unintended decision rules~\citep{sagawa2020investigation}. 
Accordingly, some studies~\citep{nam2020learning,kim2021learning} attempt to identify and emphasize the bias-conflicting samples. Nevertheless, we find that the debiasing effect is hampered by the low identification accuracy and the suboptimal emphasizing strategies. 
In this work, we build an enhanced two-stage debiasing scheme to combat unknown dataset biases. We present an Effective bias-Conflicting Scoring (ECS) function to mine bias-conflicting samples in stage \uppercase\expandafter{\romannumeral1}. On top of the off-the-shelf method, we propose a peer-picking mechanism to consciously pursue seriously biased auxiliary models and employ epoch-ensemble to obtain more accurate and stable scores.
In stage \uppercase\expandafter{\romannumeral2}, we propose Gradient Alignment (GA), which balances the gradient contributions across the mined bias-aligned and bias-conflicting samples to prevent models from being biased. In order to achieve dynamic balance throughout optimization, the gradient information is served as an indicator to down-weight (up-weight) the mined bias-aligned (bias-conflicting) samples.
Furthermore, to avoid the models relying solely on simple shortcuts to accomplish the learning objective, we introduce Self-Supervised (SS) pretext tasks in stage \uppercase\expandafter{\romannumeral2}, encouraging richer features to be considered when making decisions.
Figure~\ref{fig:intro_effect} depicts the effects of ECS, GA, and SS.

\begin{figure*}[t]
\centering
\includegraphics[width=0.95\textwidth]{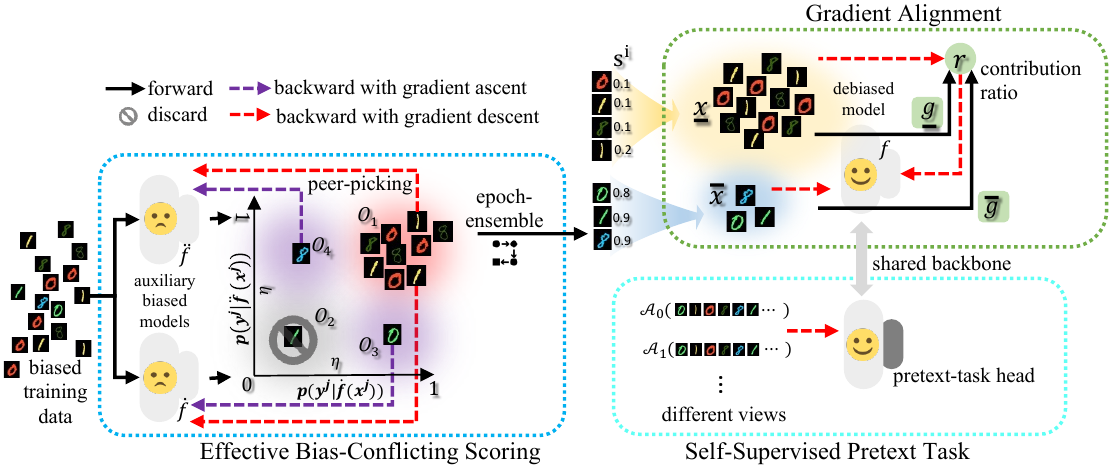}
\caption{Our debiasing scheme. \textbf{Stage \uppercase\expandafter{\romannumeral1}:} training auxiliary biased models $\dot{f}, \ddot{f}$ with peer-picking and epoch-ensemble to score the likelihood that a sample is bias-conflicting (in Section~\ref{sec:det}). \textbf{Stage \uppercase\expandafter{\romannumeral2}:} learning debiased model $f$ with gradient alignment (in Section~\ref{sec:ga}) and self-supervised pretext tasks (in Section~\ref{sec:ss}). A dashed arrow starting from a sample cluster indicates that the model is updated with gradients from these samples.
}
\label{fig:intro}
\end{figure*}

In comparison to other debiasing techniques, the proposed solution 
(i) does not rely on comprehensive bias annotations~\citep{tartaglione2021end,zhu2021learning,li2019repair,Sagawa*2020Distributionally,goel2021model,kim2019learning} or a pre-defined bias type~\citep{bahng2020learning,clark2019don,UtamaDebias2020,geirhos2018imagenet,wang2018learning}; 
(ii) does not require disentangled representations~\citep{tartaglione2021end,kim2021learning,kim2021biaswap,bahng2020learning}, which may fail in complex scenarios where disentangled features are hard to extract; 
(iii) does not introduce heavy data augmentations~\citep{geirhos2018imagenet,kim2021biaswap,kim2021learning,goel2021model}, avoiding additional training complexity such as in generative models; 
(iv) does not involve modification of model backbones~\citep{kim2021learning}, making it easy to be applied to other networks. 
(v) significantly improves the debiasing performance.
The main contributions of this work are summarized as follows:
\begin{enumerate}[(1)]
\item To combat unknown dataset biases, we present an enhanced two-stage approach (illustrated in Figure~\ref{fig:intro}) in which an effective bias-conflicting scoring algorithm equipped with peer-picking and epoch-ensemble in stage \uppercase\expandafter{\romannumeral1} (in Section~\ref{sec:det}), and gradient alignment in stage \uppercase\expandafter{\romannumeral2} (in Section~\ref{sec:ga}) are proposed.
\item In stage  \uppercase\expandafter{\romannumeral2} (in Section~\ref{sec:ss}), we introduce self-supervised pretext tasks to demonstrate the ability of the unsupervised learning paradigm to alleviate bias in supervised learning.
\item Broad experiments on commonly used datasets are conducted to compare several debiasing methods in a fair manner (overall, we train more than 700 models), among which the proposed method achieves state-of-the-art performance (in Section~\ref{sec:exp}).
\item We undertake comprehensive analysis (in Section~\ref{sec:further_ana}), including the efficacy of each component, the solution's effectiveness in various scenarios, the sensitivity of the hyper-parameters, and so on.
\end{enumerate}
\vspace{1mm}

A preliminary version of this work has been accepted by a conference~\citep{zhao2023debias}, but we extend this work with the following additions: 
(i) we further introduce self-supervised pretext tasks to help the models leverage abundant features and investigate their effectiveness with extended experiments (in Section~\ref{sec:ss} and Section~\ref{sec:quan_com});
(ii) a more detailed description and analysis of the datasets and the compared methods are provided (in Section~\ref{sec:datasets} and Section~\ref{sec:com_methods});
(iii) we present and analyze the results measured on the bias-aligned and bias-conflicting test samples separately (in Section~\ref{sec:quan_com});
(iv) we include more detailed results, such as the performance of the last epoch (in Table~\ref{tab:last_comp}), the precision-recall curves of different bias-conflicting scoring strategies (in Figure~\ref{fig:pr_curves}), the precision and recall of our mined bias-conflicting samples (in Table~\ref{tab:complete_pr}), the final debiasing results of GA with different bias-conflicting scoring methods (in Table~\ref{tab:bga_diff_mining});
(v) the analysis and discussion are extended, such as the number of auxiliary biased models (in Section~\ref{sec:hyperparameters}), when there are only a few bias-conflicting samples (in Section~\ref{app:rho_analysis}), when the training data is unbiased (in Section~\ref{app:safe}), the connection to curriculum learning (in Section~\ref{app:curriculum});
(vi) the limitation and future work are further discussed (in Section~\ref{sec:discussion}).

\section{Related work}
\label{sec:related}

\textbf{Combating biases with known types and labels.} Many debiasing approaches require explicit bias types and bias labels for each training sample. A large group of strategies aims at disentangling spurious and intrinsic features~\citep{moyer2018invariant}. For example, EnD~\citep{tartaglione2021end} designs regularizers to disentangle representations with the same bias label and entangle features with the same target label; BiasCon~\citep{hong2021unbiased} pulls samples with the same target label but different bias labels closer in the feature space based on contrastive learning; and some other studies learn disentangled representation by mutual information minimization~\citep{zhu2021learning,kim2019learning,ragonesi2021learning}. Another classic approach is to reweigh/resample training samples based on sample number or loss of different explicit groups~\citep{li2018resound,sagawa2020investigation,li2019repair}, or even to synthesize samples~\citep{agarwal2020towards}. Besides, \cite{Sagawa*2020Distributionally} and \cite{goel2021model} intend to improve the worst-group performance through group distributionally robust optimization~\citep{goh2010distributionally} and Cycle-GAN~\citep{zhu2017unpaired} based data augmentation, respectively. Furthermore, IRM~\citep{arjovsky2019invariant} is designed to learn a representation that performs well in all environments; domain-independent classifiers are introduced by~\cite{wang2020towards} to accomplish target tasks in each known bias situation.

\textbf{Combating biases with known types.} To alleviate expensive bias annotation costs, some bias-tailored methods relax the demands by requiring only the bias types~\citep{geirhos2018imagenet}. \cite{bahng2020learning} elaborately design specific networks based on the bias types to obtain biased representations on purpose (\textit{e.g.}, using 2D CNNs to extract static bias in action recognition). Then, the debiased representation is learned by encouraging it to be independent of the biased one. \cite{wang2018learning} try to project the model's representation onto the subspace orthogonal to the texture-biased representation. SoftCon~\citep{hong2021unbiased} serves as an extension of BiasCon to handle cases where only the bias type is available. In addition, the ensemble approach that consists of a bias-type customized biased model and a debiased model is employed in natural language processing as well~\citep{he2019unlearn,clark2019don,cadene2019rubi,UtamaDebias2020,clark2020learning}.

\textbf{Combating unknown biases.} Despite the effectiveness of the methodologies described above, the assumptions limit their applications, as manually discovering bias types heavily relies on experts' knowledge and labeling bias attributes for each training sample is even more laborious. As a result, recent studies~\citep{le2020adversarial,kim2019multiaccuracy,hashimoto2018fairness} try to obtain debiased models with unknown biases, which are more realistic. \cite{nam2020learning} mine bias-conflicting samples with generalized cross entropy (GCE) loss~\citep{zhang2018generalized} and emphasize them by using a designed weight assignment function. \cite{kim2021learning} further synthesize diverse bias-conflicting samples via feature-level data augmentation, whereas \cite{kim2021biaswap} directly generate them with SwapAE~\citep{park2020swapping}. RNF~\citep{du2021fairness} uses the neutralized representations from samples with the same target label but different bias labels (generated by GCE-based biased models, the version that accesses real bias labels is called RNF-GT) to train the classification head alone. Besides GCE loss, feature clustering~\citep{sohoni2020no}, early-stopping~\citep{liu2021just}, forgettable examples~\citep{yaghoobzadeh2021increasing} and limited network capacity~\citep{sanh2020learning,Utama2020TowardsDN} are involved to identify bias-conflicting samples. Furthermore,~\cite{creager2021environment} and~\cite{lahoti2020fairness} alternatively infer dataset partitions and enhance domain-invariant feature learning by min-max adversarial training. In addition to the identify-emphasize paradigm,~\cite{pezeshki2020gradient} introduces a novel regularization method for decoupling feature learning dynamics in order to improve model robustness.

\begin{algorithm*}[t]
\caption{\textbf{E}ffective bias-\textbf{C}onflicting \textbf{S}coring (ECS)}
\label{alg:det}
\KwIn{$\mathcal{D}$=$\{(x^{i},y^{i})\}_{i=1}^N$; initial models $\dot{f}^0$, $\ddot{f}^0$ and b-c scores $\{s^i\gets 0 \}_{i=1}^N$; loss function $\ell$; threshold $\eta$.}
\For{$t=0$ \KwTo $T-1$}{
$\mathcal{B} = \{(x^{j}, y^{j})\}_{j=1}^B \gets \text{FetchBatch}(\mathcal{D})$ \tcp{batch size $B$}

$ \{ p( y^{j}  \vert \dot{f}^{t}(x^{j}) ) \}, \ \{p( y^{j}  \vert \ddot{f}^{t}(x^{j}) )\} \gets \text{Forward}(\mathcal{B}, \dot{f}^{t}, \ddot{f}^{t})$ \; 

$\dot{l}^t \gets 0; \quad \ddot{l}^t \gets 0;$ \tcp{initialize loss}

\For{$j=1$ \KwTo $B$}{
\uIf{$p(y^{j}  \vert \dot{f}(x^{j})) >\eta \enspace \text{and} \enspace p(y^{j}  \vert \ddot{f}(x^{j})) >\eta$}{
$\dot{l}^t \mathrel{+}= \ell(\dot{f}^{t}(x^{j}), y^{j}); \quad \ddot{l}^t \mathrel{+}= \ell(\ddot{f}^{t}(x^{j}), y^{j})$ \; 
}
\uElseIf{$p(y^{j}  \vert \dot{f}(x^{j})) >\eta \enspace \text{and} \enspace p(y^{j}  \vert \ddot{f}(x^{j})) \leq\eta$}{
$\dot{l}^t \mathrel{-}= \ell(\dot{f}^{t}(x^{j}), y^{j})$ \; 
}
\ElseIf{$p(y^{j}  \vert \dot{f}(x^{j})) \leq\eta \enspace \text{and} \enspace p(y^{j}  \vert \ddot{f}(x^{j})) >\eta$}{
$\ddot{l}^t \mathrel{-}= \ell(\ddot{f}^{t}(x^{j}), y^{j})$ \; 
}
}
$\dot{f}^{t+1} \gets \text{Backward\&Update} (\dot{f}^{t}, \frac{\dot{l}^t}{B})$  ;\ \quad
$\ddot{f}^{t+1} \gets \text{Backward\&Update} (\ddot{f}^{t}, \frac{\ddot{l}^t}{B})$ \;

\If{$(t+1) \% T' =0$}{
\For{$i=1$ \KwTo $N$}{
$s^{i} \mathrel{+}= \frac{T'}{T} [1-\frac{p(y^{i} \vert \dot{f}^{t+1}(x^{i}))+p(y^{i} \vert \ddot{f}^{t+1}(x^{i}))}{2}]$ 
}
}
}
\KwOut{the estimated b-c scores $\{s^i\}_{i=1}^N$.}
\end{algorithm*}

\textbf{Self-supervised learning.} In recent years, self-supervised learning has achieved significant success in vision tasks. 
For applications, self-supervised learning has been employed in object recognition/detection/segmentation~\citep{he2020momentum}, video tasks~\citep{tong2022videomae}, few-shot learning~\citep{gidaris2019boosting}, manipulation detection~\citep{zeng2022towards}, \textit{etc}. 
For pretext tasks in self-supervised training, position prediction~\citep{doersch2015unsupervised}, Jigsaw puzzles~\citep{noroozi2016unsupervised}, rotation prediction~\citep{gidaris2018unsupervised}, clustering~\citep{van2020scan,caron2020unsupervised}, contrastive learning~\citep{chen2020simple,he2020momentum}, mask and reconstruct~\citep{he2022masked}, \textit{etc.} are adopted to extract transferable representations from the unlabeled data.
For the training data, besides learning on unlabeled data, self-supervised learning has also been utilized to pursue more general features with labeled~\citep{khosla2020supervised}, partial labeled~\citep{wang2022towards} or mixed data~\citep{zhai2019s4l}.

\section{Methodology}
\label{sec:method}

The whole debiasing solution is illustrated in Figure~\ref{fig:intro}. We present peer-picking, epoch-ensemble for stage \uppercase\expandafter{\romannumeral1} (in Section~\ref{sec:det}), gradient alignment and self-supervised pretext tasks for stage \uppercase\expandafter{\romannumeral2} (in Section~\ref{sec:ga} and Section~\ref{sec:ss}, respectively).

\subsection{Effective bias-conflicting scoring}
\label{sec:det}

Due to the explicit bias information is not available, we try to describe how likely input $x$ is a bias-conflicting sample via the \textbf{bias-conflicting (b-c) score}: $s(x,y)$ $\in$ [0,1], where $y \in \{1,2,\cdots,C\}$ stands for the target label. A larger $s(x,y)$ indicates that $x$ is harder to be recognized via unintended decision rules. 
As models are prone to fitting shortcuts, previous studies~\citep{kim2021biaswap,liu2021just} resort model's output probability on target class to define $s(x,y)$ as $
1 - p(y \vert \dot{f}(x) ),
$ where $
p(c\vert \dot{f}(x)) = \frac{e^{\dot{f}(x)[c]}}{\sum_{c'=1}^C e^{\dot{f}(x)[c']}}
$, $\dot{f}$ is an auxiliary biased model and $\dot{f}(x)[c]$ denotes the $c^{\text{th}}$ index of logits $\dot{f}(x)$.
Despite this, over-parameterized networks tend to ``memorize" all samples, resulting in low scores for the real bias-conflicting samples as well. To avoid it, we propose the following two strategies. The whole scoring framework is summarized in Algorithm~\ref{alg:det} (noting that the ``for'' loop is used for better clarification, which can be avoided in practice).

\textbf{Training auxiliary biased models with peer-picking.} Deliberately amplifying the auxiliary model's bias seems to be a promising strategy for better scoring~\citep{nam2020learning}, as heavily biased models can assign high b-c scores to bias-conflicting samples. We achieve this by \textbf{confident-picking} --- only picking samples with confident predictions (which are more like bias-aligned samples) to update auxiliary models. Nonetheless, a few bias-conflicting samples can still be overfitted and the memorization will be strengthened with continuous training. Thus, with the assist of \textbf{peer model}, we propose \textbf{peer-picking}, a co-training-like~\citep{han2018co} paradigm, to train auxiliary biased models.

Our method maintains two auxiliary biased models $\dot{f}$ and $\ddot{f}$ simultaneously (identical structure here). Considering a training set $\mathcal{D}$ = $\{(x^{i},y^{i})\}^N_{i=1}$ with $B$ samples in each batch, with a threshold $\eta \in (0,1)$, each model divides samples into confident and unconfident groups relying on the output probabilities on target classes. Consequently, four clusters are formed as shown in Figure~\ref{fig:intro}. For the red cluster ($\mathcal{O}_1$), since two models are confident on them, it is reasonable to believe that they are indeed bias-aligned samples, therefore we pick up them to update model via gradient descent as usual (Line 7,12 of Algorithm~\ref{alg:det}). While the gray cluster ($\mathcal{O}_2$), on which both two models are unconfident, will be discarded outright as they might be bias-conflicting samples. The remaining purple clusters ($\mathcal{O}_3$ and $\mathcal{O}_4$) indicate that some samples may be bias-conflicting, but they are memorized by one of auxiliary models. Inspired by the work for handling noisy labels~\citep{pmlr-v119-han20c}, we endeavor to force the corresponding model to forget the memorized suspicious samples via gradient ascent (Line 9,11,12). We average the output results of the two heavily biased models $\dot{f}$ and $\ddot{f}$ to obtain b-c scores (Line 15).

\textbf{Collecting results with epoch-ensemble.} During the early stage of training, b-c scores $\{s^i\}$ ($s^i$:=$s(x^{i}, y^{i})$) of real bias-conflicting samples are usually higher than those of bias-aligned ones, while the scores may be indistinguishable at the end of training due to overfitting. Unfortunately, selecting an optimal moment for scoring is strenuous. To avoid tedious hyper-parameter tuning, we collect results every $T'$ iterations (typically every epoch in practice, \textit{i.e.}, $T'=\lfloor \frac{N}{B} \rfloor$) and adopt the ensemble averages of multiple results as the final b-c scores (Line 15). We find that the ensemble can alleviate the randomness of a specific checkpoint and achieve superior results without using tricks like early-stopping.

\subsection{Gradients alignment}
\label{sec:ga}

\begin{figure}[t]
\centering
\includegraphics[width=0.8\columnwidth]{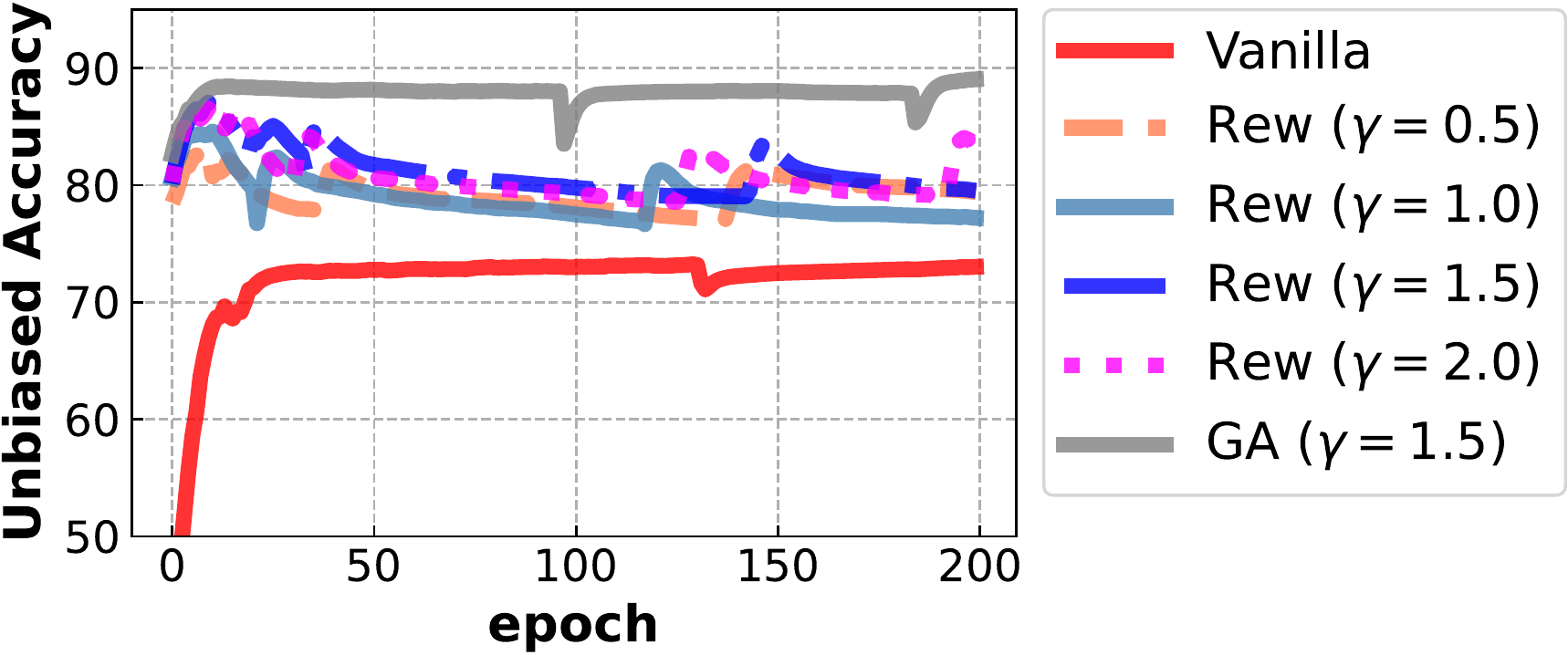} 
\caption{Unbiased accuracy on Colored MNIST.}
\label{fig:acc_c_mnist}
\end{figure}

Then, we attempt to train the debiased model $f$. We focus on an important precondition of the presence of biased models: the training objective can be achieved through unintended decision rules. To avoid it, one should develop a new learning objective that cannot be accomplished by these rules. The most straightforward inspiration is the use of plain reweighting (Rew) to intentionally rebalance sample contributions from different domains~\citep{sagawa2020investigation}:
\begin{small}
\begin{equation}
\mathcal{L}_{Rew} = \sum_{i=1}^{\underline{N}} \frac{\overline{N}}{\gamma \cdot \underline{N}} \cdot \ell(f(\underline{x}^{i}), y^{i}) + \sum_{j=1}^{\overline{N}} \ell(f(\overline{x}^{j}), y^{j}),
\label{eq:reweight}
\end{equation}
\end{small}where $\overline{N}$ and $\underline{N}$ are the number of bias-conflicting and bias-aligned samples respectively, $\gamma \in (0, \infty)$ is a reserved hyper-parameter to conveniently adjust the tendency: when $\gamma \rightarrow 0$, models intend to exploit bias-aligned samples more and when $\gamma \rightarrow \infty$, the behavior is reversed. As depicted in Figure~\ref{fig:acc_c_mnist}, assisted with Rew, unbiased accuracy skyrockets in the beginning, indicating that the model tends to learn intrinsic features in the first few epochs, while declines gradually, manifesting that the model is biased progressively (adjusting $\gamma$ can not reverse the tendency).

The above results show that the static ratio between $\overline{N}$ and $\underline{N}$ is not a good indicator to show how balanced the training is, as the influence of samples can fluctuate during training. Accordingly, we are inspired to directly choose gradient statistics as a metric to indicate whether the training is overwhelmed by bias-aligned samples. Let us revisit the commonly used cross-entropy loss:
\begin{equation}
\ell(f(x), y) = -\sum_{c=1}^{C} \mathbb{I}_{c=y} \log p(c  \vert f(x)).
\end{equation}
For a sample $(x,y)$, the gradient on logits $f(x)$ is given by 
\begin{equation}
\begin{aligned}
&\nabla_{f(x)}  \ell(f(x), y) =\\ 
&[
\frac{\partial \ell(f(x), y)}{\partial f(x)[1]},
\frac{\partial \ell(f(x), y)}{\partial f(x)[2]},
\cdots,
\frac{\partial \ell(f(x), y)}{\partial f(x)[C]}
]^{\mathsf{T}}.
\end{aligned}
\end{equation}
We define the current gradient contribution of sample $(x,y)$ as 
\begin{equation}
\begin{aligned}
g(x,y \vert f) &= \parallel \nabla_{f(x)}  \ell(f(x), y) \parallel_1\\
&=\sum_{c=1}^C  \vert \frac{\partial \ell(f(x), y)}{\partial f(x)[c]}  \vert\\ &= 2 \vert \frac{\partial \ell(f(x), y)}{\partial f(x)[y]}  \vert = 2 - 2p(y  \vert f(x)).
\end{aligned}
\end{equation} Assuming within the $t^{\text{th}}$ iteration ($t \in [0, T-1]$), the batch is composed of $\underline{B}^t$ bias-aligned and $\overline{B}^t$ bias-conflicting samples ($B$ in total, $\underline{B}^t \gg \overline{B}^t$ under our concerned circumstance). The accumulated gradient contributions generated by bias-aligned samples are denoted as 

\begin{equation}
\underline{g}^t = \sum_{i=1}^{\underline{B}^t} g(\underline{x}^i,y^i  \vert f^t),
\end{equation} similarly for the contributions of bias-conflicting samples: $\overline{g}^t$.

\begin{figure}[t]
\centering
\includegraphics[width=0.47\textwidth]{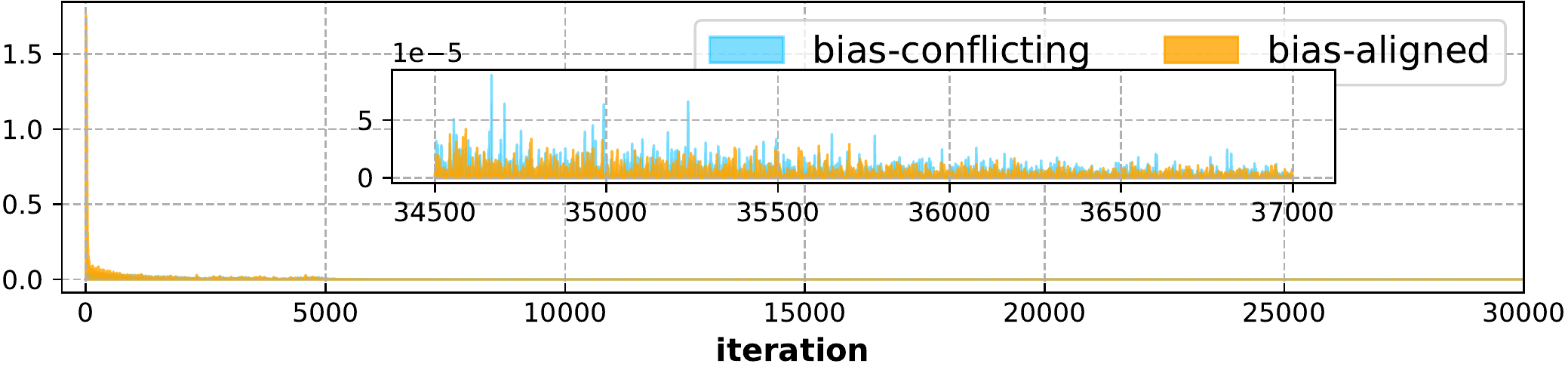}\\
\includegraphics[width=0.47\textwidth]{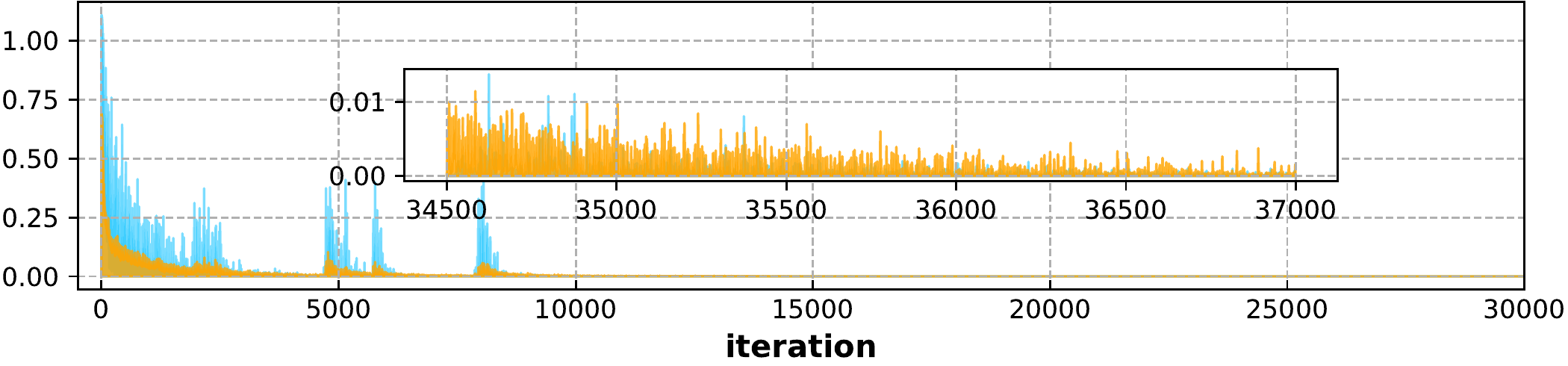}\\
\includegraphics[width=0.47\textwidth]{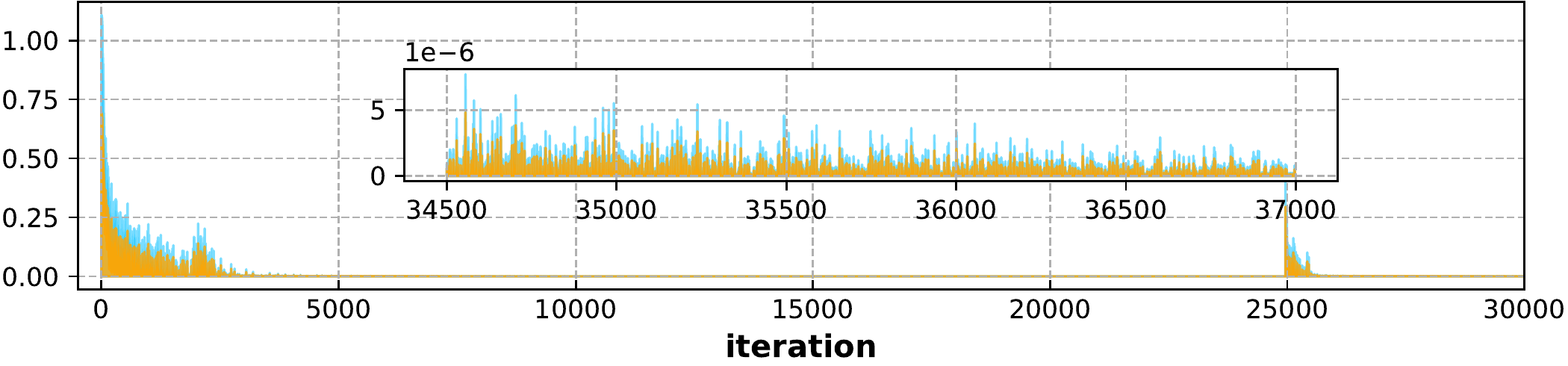}
\caption{Statistics of $\{ \overline{g}^t \}_{t=0}^{T-1}$ and $\{ \underline{g}^t \}_{t=0}^{T-1}$. Vanilla (top), Rew (middle), GA (bottom). Results in the late stage are enlarged and shown in each figure.}
\label{fig:grad_c_mnist}
\end{figure}

We present the statistics of $\{ \overline{g}^t \}_{t=0}^{T-1}$ and $\{ \underline{g}^t \}_{t=0}^{T-1}$ when learning with the standard ERM learning objective (Vanilla) and Equation~\eqref{eq:reweight} (Rew) respectively in Figure~\ref{fig:grad_c_mnist}.
For vanilla training, we find the gradient contributions of bias-aligned samples overwhelm that of bias-conflicting samples at the beginning, thus the model becomes biased towards spurious correlations rapidly. Even though at the late stage, the gap in gradient contributions shrinks, it is hard to rectify the already biased model.
For Rew, we find the contributions of bias-conflicting and bias-aligned samples are relatively close at the beginning (compared to those under Vanilla), thus both of them can be well learned. Nonetheless, the bias-conflicting samples are memorized soon due to their small quantity, and the gradient contributions from the bias-conflicting samples become smaller than that of the bias-aligned samples gradually, leading to biased models step by step.

The above phenomena are well consistent with the accuracy curves in Figure~\ref{fig:acc_c_mnist}, indicating that the gradient statistics can be a useful ``barometer'' to reflect the optimization process. Therefore, the core idea of gradient alignment is to rebalance bias-aligned and bias-conflicting samples according to their currently produced gradient contributions. Within the $t^{\text{th}}$ iteration, We define the contribution ratio $r^t$ as:
\begin{equation}
r^t = \frac{ \overline{g}^t}
{\gamma \cdot \underline{g}^t} = 
\frac{\sum_{j=1}^{ \overline{B}^t } [1 - p(y^j  \vert f^{t}(\overline{x}^{j}) ) ]}
{\gamma \cdot \sum_{i=1}^{\underline{B}^t} [1 - p(y^i  \vert f^{t}(\underline{x}^{i}) ) ]},
\label{eq:r_ga}
\end{equation}
where $\gamma$ plays a similar role as in Rew. Then, with $r^t$, we rescale the gradient contributions derived from bias-aligned samples to achieve alignment with that from bias-conflicting ones, which can be simply implemented by reweighting the learning objective for the $t^{\text{th}}$ iteration:
\begin{equation}
\mathcal{L}_{GA}^t = \sum_{i=1}^{\underline{B}^t} r^t \cdot  \ell(f^{t}(\underline{x}^{i}), y^{i}) + \sum_{j=1}^{\overline{B}^t} \ell(f^{t}(\overline{x}^{j}), y^{j}),
\label{eq:GA}
\end{equation}
\textit{i.e.}, the modulation weight is adaptively calibrated in each iteration. As shown in Equation~\eqref{eq:r_ga} and~\eqref{eq:GA}, GA only needs negligible computational extra cost (1$\times$ forward and backward as usual, only increases the cost of computing $r^t$). As shown in Figure~\ref{fig:grad_c_mnist}, GA can dynamically balance the contributions throughout the whole training process. Correspondingly, it obtains optimal and stable predictions as demonstrated in Figure~\ref{fig:acc_c_mnist} and multiple other challenging datasets in Section~\ref{sec:exp}. Noting that as bias-conflicting samples are exceedingly scarce, it is unrealistic to ensure that every class can be sampled in one batch, thus all classes share the same ratio in our design.

\begin{figure*}[t]
\centering
\includegraphics[width=0.95\textwidth]{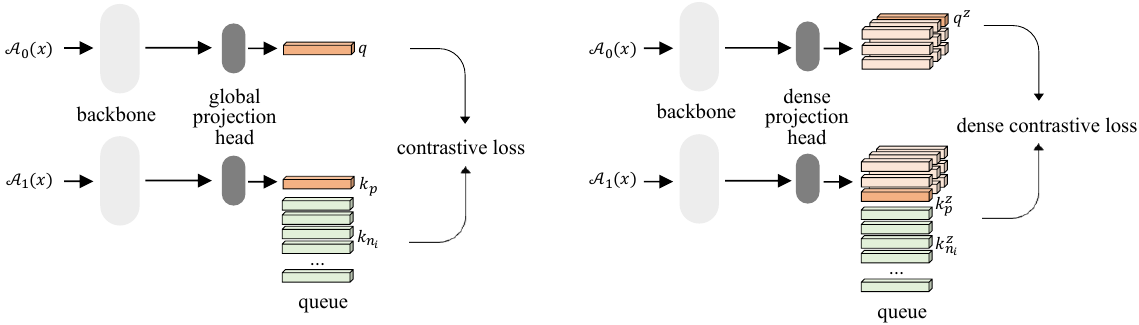} 
\caption{The illustrations of contrastive learning (left) and dense contrastive learning (right).}
\label{fig:dcl}
\end{figure*}

To handle unknown biases, we simply utilize the estimated b-c score $\{s^i\}_{i=1}^N$ and a threshold $\tau$ to assign input $x$ as bias-conflicting ($s(x,y)$ $\geq$ $\tau$) or bias-aligned ($s(x,y)$ \textless $\tau$) here. For clarity, GA with the pseudo annotations (bias-conflicting or bias-aligned) produced by ECS will be denoted as `ECS+GA' (similarly, `ECS+$\triangle$' represents combining ECS with method $\triangle$).

\subsection{Self-supervised pretext tasks}
\label{sec:ss}

The skewed feature representation is an important factor for the biased model. So, if we can help the model learn richer representation, the bias can be alleviated to some extent.
Self-supervised learning has received a lot of attention and made significant progress in recent years, allowing the model to learn transferable feature representations based on various image regions.
Inspired by the desiderata of debiasing and the ability of self-supervised learning, in this work, we investigate the efficacy of self-supervised learning on labeled data for debiasing. Specifically, we further exploit self-supervision as an auxiliary task in the debiased training scheme to pursue unbiased representations. 
The workflow is illustrated in Figure~\ref{fig:intro}.
As examples, in this work, we employ the dense contrastive learning~\citep{wang2021dense} and the rotation prediction task~\citep{gidaris2018unsupervised} as the pretext tasks. We detail the two tasks below. Other advanced self-supervision techniques can be incorporated into the pipeline similarly.

\begin{figure*}[t]
\centering
\subcaptionbox{Colored MNIST\label{fig:eg_mnist_app}}{
\includegraphics[height=0.12\textwidth]{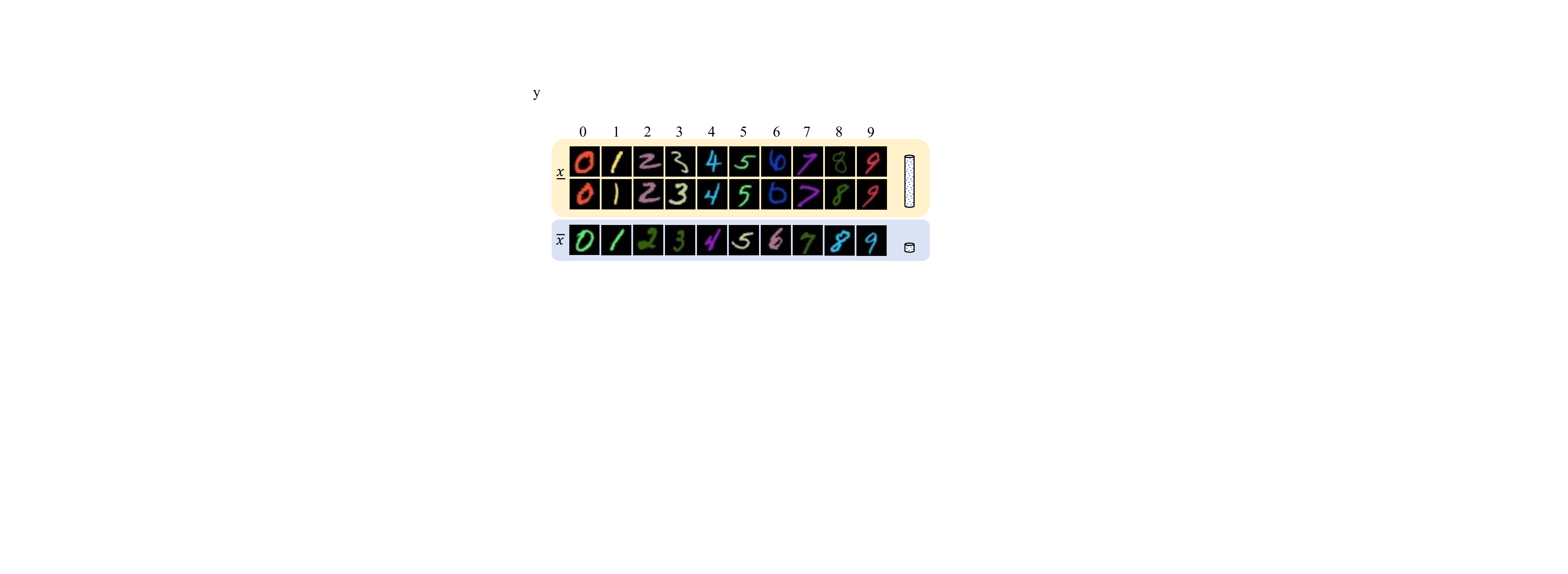}
}  \hspace{2mm}
\subcaptionbox{Corrupted CIFAR10$^1$\label{fig:eg_cifar}}{
\includegraphics[height=0.14\textwidth]{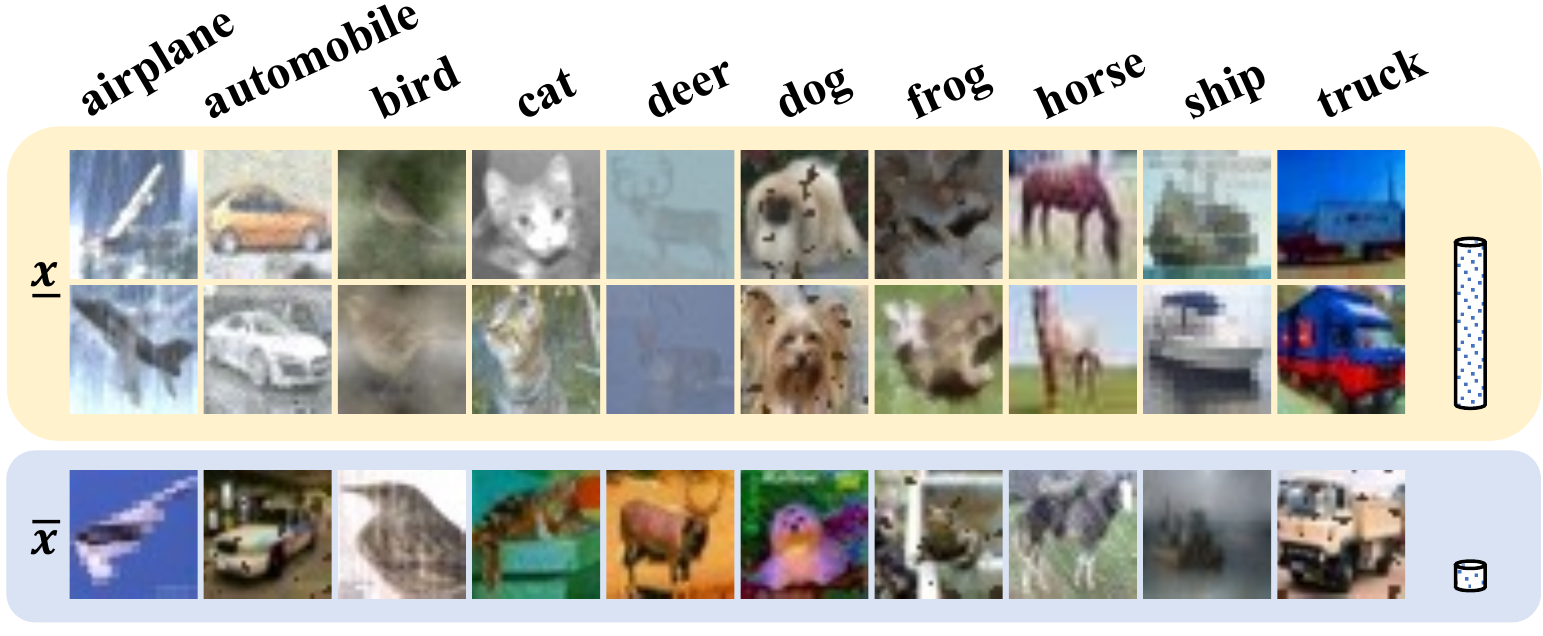}
}  \\
\subcaptionbox{Biased Waterbirds\label{fig:eg_bird}}{
\includegraphics[height=0.14\textwidth]{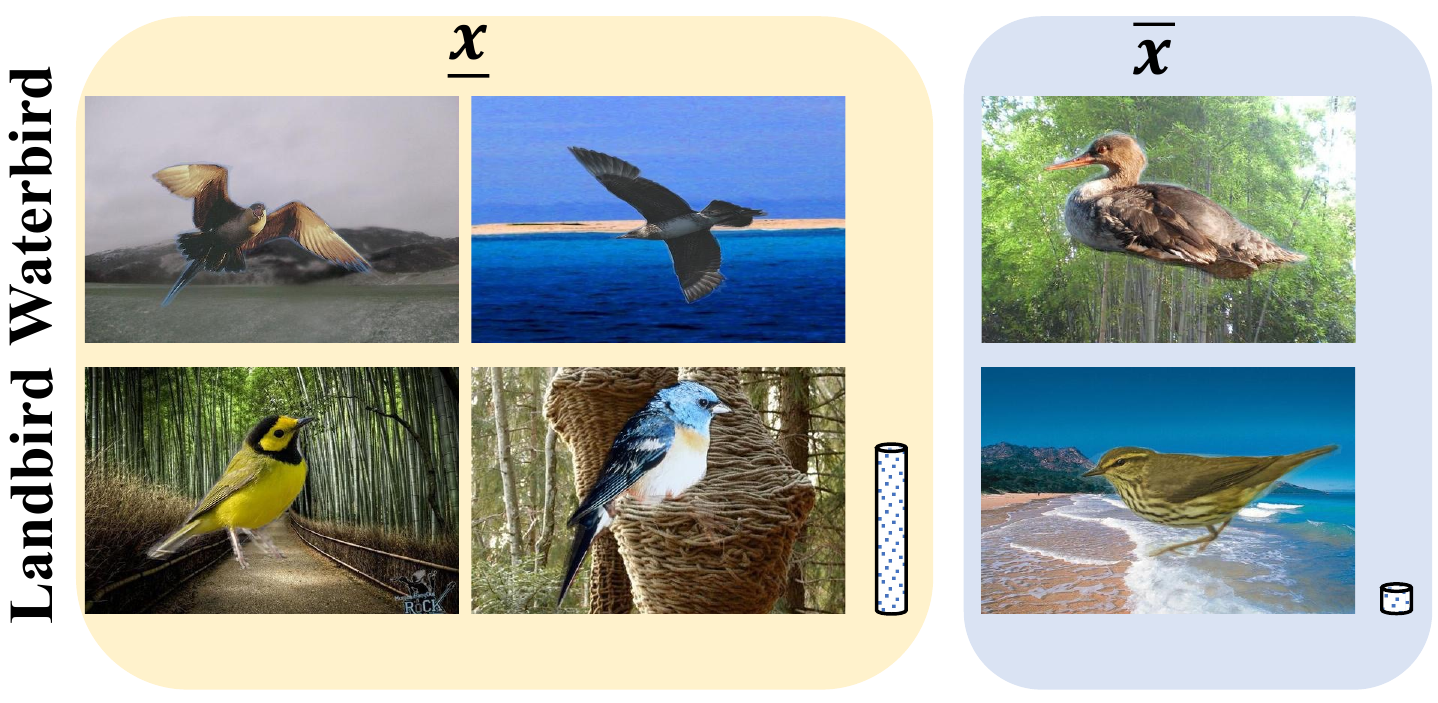}
} \hspace{2mm}
\subcaptionbox{Biased CelebA\label{fig:eg_celeba}}{
\includegraphics[height=0.14\textwidth]{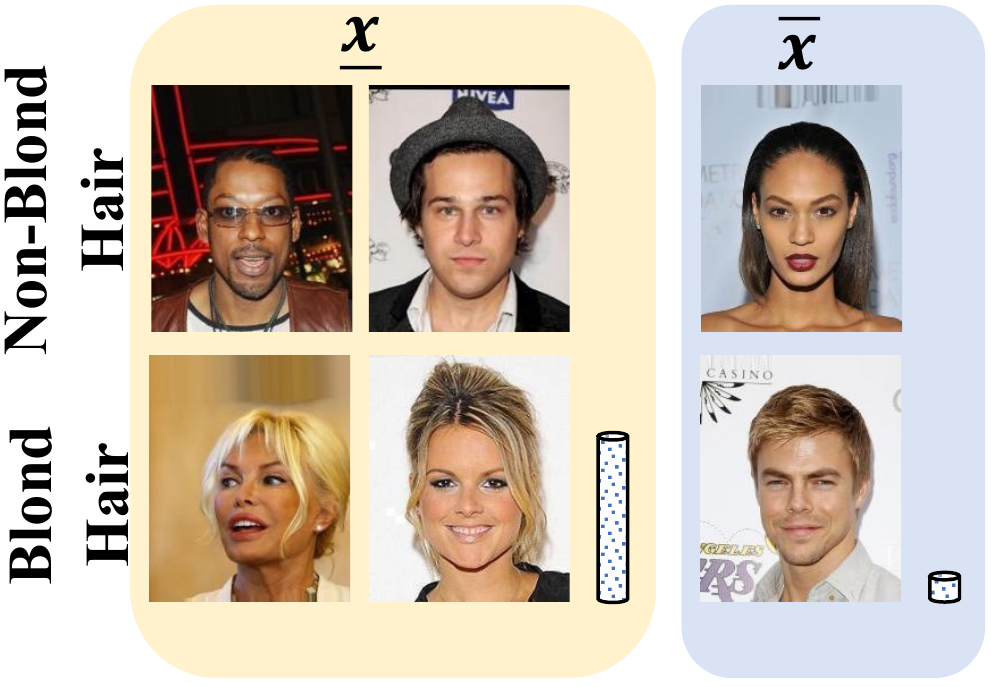}
}  \hspace{2mm}
\subcaptionbox{Multi-Color MNIST\label{fig:example_multi_bias}}{
\includegraphics[height=0.14\textwidth]{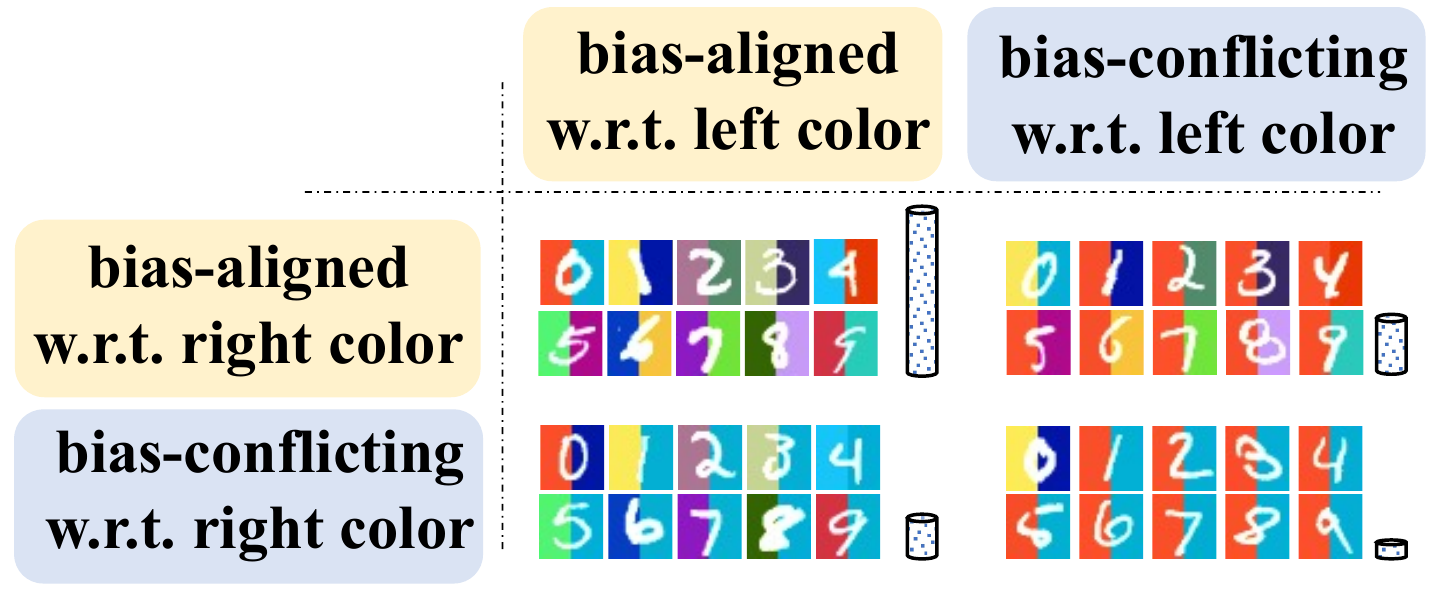}
}
\caption{Training examples. The height of the cylinder reflects the number of samples, \textit{i.e.}, most training samples are bias-aligned.}
\label{fig:examples}
\end{figure*}

\textbf{Dense contrastive learning.} Contrastive learning has achieved considerable success in self-supervised learning, which encourages the features of the positive pair to be close while pushing the representations of the negative pair away. The positive pair is typically formed by the two augmentation views ($\mathcal{A}_0$ and $\mathcal{A}_1$, $\mathcal{A}_a$ stands for the augmentations, here random crop and horizontal flip are employed) of the same image.
Following MoCo~\citep{he2020momentum}, contrastive learning can be considered as training an encoder for a dictionary look-up task as shown in Figure~\ref{fig:dcl} (left). For an encoded query $q$ (derived from $\mathcal{A}_0(x)$) and its positive key $k_p$ (derived from $\mathcal{A}_1(x)$), negative keys $\{k_{n_1}, k_{n_2}, \cdots\}$ (maintained in the queue), the contrastive learning loss is formed as:
\begin{equation}
\ell_{cl} = -\log \frac{e^{q \cdot k_p}}{e^{q \cdot k_p} + \sum_{i} e^{q \cdot k_{n_i}}}.
\end{equation}
We omit the temperature here for brevity. Commonly, the query and keys are encoded at the level of global feature. To compel the model to use richer features, we adopt a ``dense" version~\citep{wang2021dense} here which considers a dense pairwise contrastive learning task (at the level of local feature) instead of the global image classification. By replacing the global projection head with the dense projection head as depicted in Figure~\ref{fig:dcl} (right), we can obtain a $Z \times Z$ dense feature map. The $z^{\text{th}}$ query out of $Z^2$ encoded queries is denoted as $q^z$, its positive key is denoted as $k^z_p$ and a negative key is denoted as $k^z_{n_i}$, then the dense contrastive learning loss is formed as:
\begin{equation}
\ell_{dcl} = \frac{1}{Z^2} \sum_{z=1}^{Z^2} -\log \frac{e^{q^z \cdot k^z_p}}{e^{q^z \cdot k^z_p} + \sum_{i} e^{q^z \cdot k^z_{n_i}}}.
\end{equation}

The pair construction of dense contrastive learning follows~\cite{wang2021dense} and~\cite{he2020momentum}. The negative keys are the encoded local features stored in the queue. For the positive key, considering the two views' extracted feature maps before the projection head, by downsampling (average pooling) the pre-project features to also have the shape of $Z\times Z$, a similarity matrix with dimension ${Z^2\times Z^2}$ can be calculated. Assuming the $j^{\text{th}}$ pre-project feature vector from $\mathcal{A}_1(x)$ is most similar to the $i^{\text{th}}$ pre-project feature vector from $\mathcal{A}_0(x)$. Then for the features after the projection head, we can treat the corresponding $j^{\text{th}}$ post-project feature vector from $\mathcal{A}_1(x)$ as the positive key for the $i^{\text{th}}$ post-project feature vector from $\mathcal{A}_0(x)$.

\textbf{Rotation prediction.} The main idea of image rotation prediction is to predict the rotation degree of the deliberately rotated input images, resulting a 4-class classification problem. The loss function for each sample is formulated by:
\begin{equation}
\ell_{rot} = \frac{1}{4}   \sum_{a=0}^3 
\ell(f_{rot}(\mathcal{A}_a (x)), a),
\end{equation}
here $\{\mathcal{A}_0, \mathcal{A}_1, \mathcal{A}_2, \mathcal{A}_3\}$ is the set of transformations with 4 rotation degrees $\{0^{\circ}, 90^{\circ}, 180^{\circ}, 270^{\circ}\}$, $\ell$ is the cross-entropy loss.

\section{Experiments}
\label{sec:exp}

\subsection{Datasets}
\label{sec:datasets}

We mainly conduct experiments on five benchmark datasets. Some examples from the used datasets are exhibited in Figure~\ref{fig:examples}.
For Colored MNIST (C-MNIST), the task is to recognize digits (0 - 9), in which the images of each target class are dyed by the corresponding color with probability $\rho \in \{95\%, 98\%, 99\%, 99.5\%\}$ and by other colors with probability $1-\rho$ (a higher $\rho$ indicates more severe biases).
Similarly, for Corrupted CIFAR10, each object class in it holds a spurious correlation with a corruption type. Two sets of corruption protocols are utilized, leading to two biased datasets~\citep{nam2020learning}: Corrupted CIFAR10$^1$ and CIFAR10$^2$ (C-CIFAR10$^1$, C-CIFAR10$^2$) with $\rho \in \{95\%, 98\%, 99\%, 99.5\%\}$. Following previous work~\citep{nam2020learning}, Corrupted CIFAR10$^1$ is constructed with corruption types \{\textit{Snow, Frost, Fog, Brightness, Contrast, Spatter, Elastic, JPEG, Pixelate, Saturate}\}; Corrupted CIFAR10$^2$ is constructed with corruption types \{\textit{GaussianNoise, ShotNoise, ImpulseNoise, SpeckleNoise, GaussianBlur, DefocusBlur, GlassBlur, MotionBlur, ZoomBlur, Original}\}.
In Biased Waterbirds (B-Birds)\footnote{The data is available at \url{https://nlp.stanford.edu/data/dro/waterbird_complete95_forest2water2.tar.gz}.} which is a composite dataset that superimposes foreground bird images from CUB~\citep{welinder2010caltech} onto background environment images from Places~\citep{zhou2017places}, ``waterbirds'' and ``landbirds'' are highly correlated with ``wet'' and ``dry'' habitats (95\% bias-aligned samples, \textit{i.e.}, $\rho=95\%$). Consequently, the task aiming to distinguish images as ``waterbird" or ``landbird" can be influenced by background.
In Biased CelebA (B-CelebA) which is established for face recognition where each image contains multiple attributes~\citep{liu2015deep}\footnote{The data is available at \url{http://mmlab.ie.cuhk.edu.hk/projects/CelebA.html}.}, blond hair is predominantly found in women, whereas non-blond hair mostly appears in men ($\rho=99\%$). When the goal is to classify the hair color as ``blond" or ``non-blond", the information of gender (``male'' or ``female'' in this dataset) can be served as a shortcut~\citep{nam2020learning}.
To focus on the debiasing problem, we balance the number of images per target class in B-Birds and B-CelebA.

\subsection{Compared methods}
\label{sec:com_methods}

We choose various methods for comparison: standard ERM (Vanilla), Focal loss~\citep{lin2017focal}, plain reweighting~\citep{sagawa2020investigation} (Rew and ECS+Rew), REBIAS~\citep{bahng2020learning}, BiasCon~\citep{hong2021unbiased}, RNF-GT~\citep{du2021fairness}, GEORGE~\citep{sohoni2020no}, LfF~\citep{nam2020learning}, DFA~\citep{kim2021learning}, SD~\citep{pezeshki2020gradient}, ERew~\citep{clark2019don} and PoE~\citep{clark2019don} (ECS+ERew and ECS+PoE)\footnote{As the auxiliary biased models used in ERew and PoE are designed for NLP tasks, here, we combine our ECS with them.}. Among them, REBIAS requires bias types; Rew, BiasCon, RNF-GT, and GA are performed with real bias-conflicting or bias-aligned annotations\footnote{As stated in the original papers, BiasCon and RNF-GT have variations that do not require real annotations assist with various auxiliary biased models. We only provide the upper bound of these strategies when combating unknown biases, as we found that the auxiliary models have a significant impact on the outcomes.}. We present a brief analysis of these debiasing approaches as follows based on technique categories.

\textbf{Reweighting-based strategies}. Rew is a straightforward static reweighting strategy based on the number of samples per group. Both LfF and ERew reassign sample weights assisted with a biased model but differ in weight assignment functions. ERew is also a static reweighting approach that employs output scores of a pre-trained biased model as the weight indicator. LfF applies dynamic weight adjustments during training. LfF and ERew just reweight a sample with the information from itself, whereas Rew uses global information within one minibatch to obtain sample weight\footnote{LfF is implemented at~\url{https://github.com/alinlab/LfF} and ERew is implemented at~\url{https://github.com/UKPLab/emnlp2020-debiasing-unknown}.}.

\textbf{Feature disentanglement}. REBIAS designs specific networks according to the bias type for obtaining biased representations intentionally (for our experiments, we employ CNNs with smaller receptive fields for capturing texture bias according to the original paper). Then the debiased representation is learned by encouraging it to be independent of the biased one, during which Hilbert-Schmidt Independence Criterion (HSIC) is employed to measure the degree of independence between the two representations. Building on LfF, DFA further introduces disentangled representations to augment bias-conflicting samples at the feature level. 
The methods try to explicitly extract disentangled feature representations, which is difficult to be achieved in complex datasets and tasks.
BiasCon directly uses contrastive learning to pull the same target class but different bias class sample pairs closer than the other pairs\footnote{
REBIAS is implemented at~\url{https://github.com/clovaai/rebias}. 
DFA is implemented at~\url{https://github.com/kakaoenterprise/Learning-Debiased-Disentangled}.
BiasCon is implemented at~\url{https://github.com/grayhong/bias-contrastive-learning}.
}.

\textbf{Distributionally robust optimization (DRO)}. Many previous studies resort to DRO to achieve model fairness.
GEORGE performs clustering based on the feature representations of the auxiliary biased models first and then expects to obtain fair models by using DRO with the pseudo groups. However, due to overfitting, we find that clustering with features generated from vanilla biased models is not robust and accurate, resulting in substantially inferior performance when performing DRO using the imprecise clusters\footnote{
We adopt the clustering methods utilized in GEORGE referring to~\url{https://github.com/HazyResearch/hidden-stratification}.
}.

\textbf{Ensemble approaches}. Product-of-Experts (PoE) is widely adopted in NLP-related debiasing tasks, which tries to train a debiased model in an ensemble manner with an auxiliary biased model, by combining the softmax outputs produced from the biased and debiased models\footnote{The method is implemented at~\url{https://github.com/UKPLab/emnlp2020-debiasing-unknown}.
}.

\textbf{Regularization methods}. In addition, SD directly replaces the common $l_2$ regularization with an $l_2$ penalty on the model's logits. The optimal strength of the regularization term can be hard to search, which may be very different for various datasets and tasks\footnote{It is implemented at~\url{https://github.com/mohammadpz/Gradient_Starvation}.
}.

\subsection{Evaluation metrics}

Following~\cite{nam2020learning}, we mainly report the overall unbiased accuracy, alongside the accuracy of bias-aligned and bias-conflicting test samples individually.
For experiments on Colored MNIST, Corrupted CIFAR10$^1$ and CIFAR10$^2$, we evaluate models on the unbiased test sets in which the bias attributes are independent of the target labels. For Biased Waterbirds and CelebA, to evaluate unbiased accuracy with the official test sets which are biased and imbalanced, the accuracies of each (\textit{target}, \textit{bias}) group are calculated separately and then averaged to generate the overall accuracy~\citep{nam2020learning}.

We also show the fairness performance in terms of DP and EqOdd~\citep{reddy2021benchmarking}.
For the definitions of DP and EqOdd, following~\citet{reddy2021benchmarking}, let $x$, $y$, $b$, $y'$ denote the input, target label, the bias label, and the model's prediction respectively, Demographic Parity (DP) is defined as $1 - \vert p(y'=1 \vert b=1) - p(y'=1 \vert b=0) \vert$; Equality of Opportunity \textit{w.r.t} $y=1$ (EqOpp1) is defined as $1 - \vert p(y' = 1 \vert y = 1, b = 0) - p(y' = 1 \vert y = 1, b = 1) \vert$ and Equality of Opportunity \textit{w.r.t} $y = 0$ (EqOpp0) is defined as $1 - \vert p(y' = 1 \vert y = 0, b = 0) - p(y' = 1 \vert y = 0, b = 1) \vert$, Equality of Odds (EqOdd) is defined as $0.5 \times$(EqOpp0 + EqOpp1).

\subsection{Implementation}

The studies for the previous debiasing approaches are usually conducted with varying network architectures and training schedules. We run the representative methods with identical configurations to make fair comparisons. We use an MLP with three hidden layers (each hidden layer comprises 100 hidden units) for C-MNIST, except for the biased models in REBIAS (using CNN). ResNet-20~\citep{he2016deep} is employed for C-CIFAR10$^1$ and C-CIFAR10$^2$. ResNet-18 is utilized for B-Birds and B-CelebA.
We implement all methods with PyTorch~\citep{paszke2019pytorch} and run them on a Tesla V100 GPU. 
For experiments on Colored MNIST, we use Adam optimizer to train models for 200 epochs with learning rate 0.001, batch size 256, without any data augmentation techniques. For Corrupted CIFAR10$^1$ and CIFAR10$^2$, models are trained for 200 epochs with Adam optimizer, learning rate 0.001, batch size 256, image augmentation including only random crop and horizontal flip. For Biased Waterbirds and CelebA, models are trained from imagenet pre-trained weights (Pytorch torchvision version) for 100 epochs with Adam optimizer, learning rate 0.0001, batch size 256, and horizontal flip augmentation technique.
Dense contrastive learning is utilized on B-Birds and rotation prediction is employed on C-CIFAR10$^1$ and C-CIFAR10$^2$ (as we find dense prediction is not suitable for images with very small resolution).
The code and README are provided in the supplementary material.

\begin{sidewaystable*}[htbp]
\centering
\begin{minipage}{0.99\textheight}
\caption{Overall unbiased accuracy (\%) and standard deviation over three runs. Best results with unknown biases are shown in bold. $^\dag$ indicates that the method requires prior knowledge regarding bias.}
\resizebox{\textwidth}{!}{
\begin{tabular}{l | c c c c | c c c c | c c c c | c | c}
\toprule
& \multicolumn{4}{c|}{Colored MNIST} & \multicolumn{4}{c|}{Corrupted CIFAR10$^1$} & \multicolumn{4}{c|}{Corrupted CIFAR10$^2$} & B-Birds  & B-CelebA \\
\multicolumn{1}{c|}{$\rho$} & 95\% & 98\% & 99\% & 99.5\% & 95\% & 98\% & 99\% & 99.5\% & 95\% & 98\% & 99\% & 99.5\% & 95\%  & 99\% \\
\midrule
Vanilla & 85.7$_{\pm 0.1}$  &    73.6$_{\pm 0.5}$   &   60.7$_{\pm 0.6}$   &    45.4$_{\pm 0.8}$   &   44.9$_{\pm 1.0}$   &  30.4$_{\pm 1.0}$   & 22.4$_{\pm 0.8}$  &  17.9$_{\pm 0.9}$   &  42.7$_{\pm 0.9}$   &   27.2$_{\pm 0.6}$  &   20.6$_{\pm 0.5}$ &  17.4$_{\pm 0.8}$ &  77.1$_{\pm 1.5}$ & 77.4$_{\pm 1.6}$\\ 
Focal &  86.7$_{\pm 0.2}$   &   75.8$_{\pm 0.6}$    & 62.4$_{\pm 0.3}$   &   45.9$_{\pm 0.9}$  & 45.5$_{\pm 1.0}$   &    
30.7$_{\pm 1.1}$ &  22.9$_{\pm 1.1}$  &  
17.8$_{\pm 0.5}$  & 41.9$_{\pm 0.5}$  &  26.9$_{\pm 0.5}$  &  21.0$_{\pm 0.6}$   &  17.0$_{\pm 0.2}$   &   78.6$_{\pm 0.7}$  &  78.1$_{\pm 1.0}$  \\ 
  GEORGE  &  87.0$_{\pm 0.5}$   &  76.2$_{\pm 0.9}$  & 62.4$_{\pm 0.6}$    &   46.4$_{\pm 0.2}$  & 44.6$_{\pm 1.0}$  & 29.5$_{\pm 1.0}$  &   21.8$_{\pm 0.3}$   &    17.9$_{\pm 0.6}$   &    44.2$_{\pm 1.9}$  &  27.3$_{\pm 1.6}$   &    20.7$_{\pm 1.2}$   &    17.7$_{\pm 0.3}$  & 79.3$_{\pm 0.9}$   & 78.2$_{\pm 0.9}$  \\ 
 LfF   &  88.2$_{\pm 0.9}$ & 86.7$_{\pm 0.6}$  &   80.3$_{\pm 1.2}$ & 73.2$_{\pm 0.9}$   &   59.6$_{\pm 0.8}$ & 50.4$_{\pm 0.5}$ &  42.9$_{\pm 2.8}$    &  34.6$_{\pm 2.3}$  &   58.5$_{\pm 0.8}$ &   49.0$_{\pm 0.4}$   &   42.2$_{\pm 1.1}$ & 33.4$_{\pm 1.2}$ & 80.4$_{\pm 1.1}$ &  84.4$_{\pm 1.5}$ \\ 
 DFA    &  89.8$_{\pm 0.2}$  &  86.9$_{\pm 0.4}$ &  81.8$_{\pm 1.1}$ & 74.1$_{\pm 0.8}$ & 58.2$_{\pm 1.8}$  & 50.0$_{\pm 2.3}$   & 41.8$_{\pm 4.7}$  &  35.6$_{\pm 4.6}$  & 58.6$_{\pm 0.2}$ & 48.7$_{\pm 0.6}$  & 41.5$_{\pm 2.2}$    & 35.2$_{\pm 1.9}$  &79.5$_{\pm 0.7}$  & 84.3$_{\pm 0.6}$ \\ 
 SD    &  86.7$_{\pm 0.3}$   &  73.9$_{\pm 0.2}$   &  59.7$_{\pm 0.5}$  &   42.4$_{\pm 1.1}$   &   43.1$_{\pm 0.5}$  &   28.6$_{\pm 1.5}$   &   21.6$_{\pm 0.9}$   &   17.7$_{\pm 0.6}$   &   41.4$_{\pm 0.3}$    &   27.0$_{\pm 0.8}$    &    20.0$_{\pm 0.2}$   &  17.5$_{\pm 0.3}$  & 76.8$_{\pm 1.3}$ &  77.8$_{\pm 1.1}$  \\ 
  ECS+Rew  &   91.8$_{\pm 0.2}$  &   88.6$_{\pm 0.7}$  &  84.2$_{\pm 0.3}$  & 78.9$_{\pm 0.9}$  & 58.5$_{\pm 0.0}$  &  47.5$_{\pm 0.6}$  &  38.6$_{\pm 1.1}$  & 33.4$_{\pm 1.2}$  &  61.4$_{\pm 0.7}$   &  53.2$_{\pm 0.3}$  &  47.4$_{\pm 1.2}$  &  40.3$_{\pm 0.6}$  & 82.7$_{\pm 0.7}$   & 88.3$_{\pm 0.4}$  \\ 
 ECS+ERew  & 91.0$_{\pm 0.2}$  &  87.5$_{\pm 0.2}$  &  81.4$_{\pm 0.9}$ &   71.3$_{\pm 2.2}$  &    59.8$_{\pm 0.5}$   &   47.9$_{\pm 1.0}$    &   38.5$_{\pm 0.2}$    &   30.2$_{\pm 1.3}$    &  62.2$_{\pm 0.5}$  &     51.1$_{\pm 0.2}$  &   41.4$_{\pm 0.9}$    &  25.9$_{\pm 1.6}$  &  84.9$_{\pm 0.9}$ & 80.5$_{\pm 0.6}$  \\ 
 ECS+PoE   &   80.2$_{\pm 1.5}$    &  75.4$_{\pm 1.4}$ & 64.4$_{\pm 2.7}$  & 50.0$_{\pm 3.0}$  & 54.4$_{\pm 0.2}$   &     48.7$_{\pm 1.3}$  &   \textbf{45.6}$_{\pm 1.3}$  &  \textbf{42.7}$_{\pm 0.8}$  & 47.9$_{\pm 0.8}$      &     40.3$_{\pm 1.3}$  &    36.8$_{\pm 2.5}$   & \textbf{42.4}$_{\pm 2.3}$ & 85.8$_{\pm 0.6}$ &  81.1$_{\pm 0.1}$ \\ 
  ECS+GA  &  \textbf{92.1}$_{\pm 0.1}$  &   \textbf{89.5}$_{\pm 0.4}$  &  \textbf{86.4}$_{\pm 0.5}$   &  \textbf{79.9}$_{\pm 0.8}$  &  \textbf{61.0}$_{\pm 0.1}$   &  \textbf{51.7}$_{\pm 0.5}$   &  42.6$_{\pm 0.7}$ &   35.0$_{\pm 0.5}$  &  \textbf{64.1}$_{\pm 0.3}$  & \textbf{57.0}$_{\pm 0.6}$    & \textbf{50.0}$_{\pm 1.5}$   & 41.8$_{\pm 0.8}$ &  \textbf{86.1}$_{\pm 0.5}$   &  \textbf{89.5}$_{\pm 0.5}$\\ 
\hline \hline
$^\dag$REBIAS &  85.5$_{\pm 0.6}$  &  74.0$_{\pm 0.7}$  &  61.1$_{\pm 0.8}$  &  44.5$_{\pm 0.4}$ & 44.8$_{\pm 0.3}$   &  29.9$_{\pm 0.7}$ &  22.4$_{\pm 1.1}$   &  17.7$_{\pm 0.3}$  &  41.5$_{\pm 1.0}$ & 27.0$_{\pm 0.6}$  & 20.6$_{\pm 0.6}$   & 17.9$_{\pm 0.3}$ & 77.5$_{\pm 0.6}$  & 78.1$_{\pm 1.2}$  \\ 
$^\dag$Rew  &   91.5$_{\pm 0.0}$  &   87.9$_{\pm 0.4}$     &  83.8$_{\pm 0.6}$  & 77.6$_{\pm 0.7}$  & 59.1$_{\pm 0.2}$  &  48.9$_{\pm 0.8}$  &  40.4$_{\pm 0.4}$  & 33.4$_{\pm 1.4}$  &  61.1$_{\pm 0.2}$   &  53.1$_{\pm 0.8}$  &  46.9$_{\pm 1.1}$  &  41.2$_{\pm 0.6}$  & 86.0$_{\pm 0.4}$   & 90.7$_{\pm 0.4}$  \\ 
$^\dag$RNF-GT &   84.3$_{\pm 4.1}$  &   75.9$_{\pm 3.6}$     &  66.3$_{\pm 8.2}$  & 59.1$_{\pm 5.7}$  & 52.1$_{\pm 0.7}$  &  39.1$_{\pm 1.2}$  &  30.6$_{\pm 1.3}$  & 22.2$_{\pm 0.4}$  &  50.3$_{\pm 1.0}$   &  34.9$_{\pm 0.5}$  &  27.9$_{\pm 0.6}$  &  19.8$_{\pm 0.4}$  & 81.2$_{\pm 1.3}$   & 85.1$_{\pm 2.7}$  \\ 
$^\dag$BiasCon  &   90.9$_{\pm 0.1}$  &   86.7$_{\pm 0.1}$     &  83.0$_{\pm 0.0}$  & 79.0$_{\pm 1.5}$  & 59.0$_{\pm 0.6}$  &  48.6$_{\pm 0.6}$  &  39.0$_{\pm 0.4}$  & 32.4$_{\pm 0.3}$  &  60.0$_{\pm 0.3}$   &  49.9$_{\pm 0.3}$  &  43.0$_{\pm 0.4}$  &  37.4$_{\pm 0.8}$  & 84.1$_{\pm 0.6}$   & 90.4$_{\pm 1.2}$  \\ 
$^\dag$GA  &  92.4$_{\pm 0.3}$ &  89.1$_{\pm 0.2}$ &  85.7$_{\pm 0.4}$ &  80.4$_{\pm 0.5}$    &  61.5$_{\pm 0.8}$   &  52.9$_{\pm 0.3}$ & 43.5$_{\pm 1.6}$  & 33.9$_{\pm 0.8}$ & 64.5$_{\pm 0.2}$  & 56.9$_{\pm 0.2}$  & 51.1$_{\pm 0.3}$ & 43.6$_{\pm 0.8}$ & 87.9$_{\pm 0.5}$  &  92.3$_{\pm 0.2}$ \\ 
\bottomrule
\end{tabular}%
}
\label{tab:overall_acc}%
\end{minipage}
\\ \vspace{2mm}
\begin{minipage}{0.99\textheight}
\caption{Overall unbiased accuracy and standard deviation of the last epoch over 3 runs (\%). Best results with unknown biases are in bold. $^\dag$ indicates that they require prior knowledge regarding biases.}
\resizebox{\textwidth}{!}{
\begin{tabular}{l | c c c c | c c c c | c c c c | c | c}
\toprule
& \multicolumn{4}{c|}{Colored MNIST} & \multicolumn{4}{c|}{Corrupted CIFAR10$^1$} & \multicolumn{4}{c|}{Corrupted CIFAR10$^2$} & B-Birds  & B-CelebA \\
\multicolumn{1}{c|}{$\rho$} & 95\% & 98\% & 99\% & 99.5\% & 95\% & 98\% & 99\% & 99.5\% & 95\% & 98\% & 99\% & 99.5\% & 95\%  & 99\% \\
\midrule
Vanilla & 85.3$_{\pm 0.2}$  &    73.5$_{\pm 0.6}$   &   59.5$_{\pm 0.6}$   &    43.2$_{\pm 1.0}$   &   42.6$_{\pm 0.4}$   &  27.7$_{\pm 1.0}$   & 19.8$_{\pm 1.0}$  &  15.6$_{\pm 0.8}$   &  39.3$_{\pm 0.6}$   &   25.3$_{\pm 1.3}$  &   18.5$_{\pm 0.5}$ &  14.2$_{\pm 0.3}$ &  75.7$_{\pm 0.8}$ & 71.0$_{\pm 1.0}$\\ 
Focal &  86.7$_{\pm 0.2}$   &   75.2$_{\pm 0.4}$    & 61.7$_{\pm 0.8}$   &   44.2$_{\pm 0.7}$  & 43.6$_{\pm 1.4}$   &    27.6$_{\pm 2.0}$ &  20.9$_{\pm 0.9}$  &  14.8$_{\pm 0.7}$  & 39.2$_{\pm 1.0}$  &  25.3$_{\pm 1.4}$  &  18.2$_{\pm 0.1}$   &  14.5$_{\pm 0.2}$   &   76.4$_{\pm 0.3}$  &  71.8$_{\pm 0.8}$  \\ 
GEORGE  &  86.7$_{\pm 0.2}$   &  74.3$_{\pm 0.8}$  & 60.1$_{\pm 0.9}$    &   43.8$_{\pm 1.2}$  & 41.3$_{\pm 1.7}$  & 27.3$_{\pm 0.3}$  &   19.1$_{\pm 0.1}$   &    14.7$_{\pm 1.0}$   &    41.4$_{\pm 1.1}$  &  25.4$_{\pm 2.4}$   &    18.5$_{\pm 1.6}$   &    14.8$_{\pm 0.9}$  & 76.3$_{\pm 0.3}$   & 70.5$_{\pm 0.4}$  \\ 
 LfF   &  78.0$_{\pm 1.8}$ & 75.1$_{\pm 0.3}$  &   68.8$_{\pm 3.1}$ & 67.8$_{\pm 1.5}$   &   56.7$_{\pm 2.1}$ & 49.4$_{\pm 0.7}$ &  39.8$_{\pm 1.9}$    &  32.1$_{\pm 2.0}$  &   57.8$_{\pm 0.8}$ &   47.3$_{\pm 0.2}$   &   40.5$_{\pm 1.4}$ & 31.3$_{\pm 1.0}$ & 76.9$_{\pm 2.0}$ &  61.0$_{\pm 1.2}$ \\ 
 DFA   &  83.6$_{\pm 0.9}$  &  81.2$_{\pm 1.9}$ &  76.0$_{\pm 3.2}$ & 65.7$_{\pm 0.8}$ & 54.8$_{\pm 0.5}$  & 47.9$_{\pm 1.8}$   & 39.5$_{\pm 5.1}$  &  33.4$_{\pm 4.0}$  & 56.5$_{\pm 0.9}$ & 46.0$_{\pm 1.2}$  & 39.1$_{\pm 2.7}$    & 33.5$_{\pm 2.6}$  &74.5$_{\pm 1.1}$  & 73.2$_{\pm 3.7}$ \\ 
 SD    &  86.3$_{\pm 0.3}$   &  73.6$_{\pm 0.3}$   &  58.4$_{\pm 0.2}$  &   39.9$_{\pm 0.9}$   &   40.5$_{\pm 0.7}$  &   25.0$_{\pm 0.3}$   &   19.3$_{\pm 1.2}$   &   14.9$_{\pm 0.2}$   &   38.5$_{\pm 0.5}$    &   23.3$_{\pm 0.2}$    &    17.8$_{\pm 0.9}$   &  14.5$_{\pm 0.1}$  & 76.0$_{\pm 0.9}$ &  70.6$_{\pm 0.4}$   \\ 
  ECS+Rew  &   87.7$_{\pm 0.2}$  &   79.6$_{\pm 0.2}$  &  67.4$_{\pm 0.7}$  & 58.3$_{\pm 6.0}$  & 56.0$_{\pm 0.4}$  &  42.6$_{\pm 1.1}$  &  34.2$_{\pm 0.5}$  & 29.4$_{\pm 1.8}$  &  56.6$_{\pm 0.4}$   &  48.8$_{\pm 1.1}$  &  39.5$_{\pm 2.8}$  &  31.1$_{\pm 1.4}$  & 77.2$_{\pm 0.6}$   & 80.5$_{\pm 5.9}$  \\ 
 ECS+ERew  & 88.8$_{\pm 0.2}$  &  81.7$_{\pm 2.6}$  &  68.5$_{\pm 3.6}$ &   51.4$_{\pm 1.5}$  &    57.2$_{\pm 1.2}$   &   44.1$_{\pm 0.3}$    &   34.6$_{\pm 1.2}$    &   24.0$_{\pm 0.6}$    &  59.7$_{\pm 0.6}$  &     45.6$_{\pm 1.0}$  &   35.2$_{\pm 2.6}$    &  20.8$_{\pm 0.7}$  &  80.8$_{\pm 2.0}$ & 74.9$_{\pm 0.4}$  \\ 
 ECS+PoE   &   80.1$_{\pm 1.5}$    &  75.0$_{\pm 1.5}$ & 64.1$_{\pm 2.5}$  & 49.1$_{\pm 2.3}$  & 52.8$_{\pm 0.6}$   &     45.3$_{\pm 2.2}$  &   40.4$_{\pm 1.0}$  &  \textbf{39.4}$_{\pm 0.9}$  & 46.7$_{\pm 1.5}$      &     37.9$_{\pm 0.3}$  &    35.7$_{\pm 1.6}$   & 39.6$_{\pm 1.8}$ & 83.2$_{\pm 2.2}$ &  74.5$_{\pm 0.1}$ \\ 
  ECS+GA   &  \textbf{91.8}$_{\pm 0.2}$  &   \textbf{88.9}$_{\pm 1.1}$  &  \textbf{84.6}$_{\pm 0.7}$   &  \textbf{78.7}$_{\pm 0.5}$  &  \textbf{59.6}$_{\pm 0.5}$   &  \textbf{50.8}$_{\pm 1.0}$   &  \textbf{40.9}$_{\pm 0.3}$ &   34.3$_{\pm 0.4}$  &  \textbf{62.2}$_{\pm 0.6}$  & \textbf{55.4}$_{\pm 2.6}$    & \textbf{49.3}$_{\pm 1.2}$   & \textbf{40.5}$_{\pm 0.5}$ &  \textbf{85.5}$_{\pm 0.9}$   &  \textbf{87.4}$_{\pm 1.8}$\\ 
\hline \hline
$^\dag$REBIAS  &  85.3$_{\pm 0.5}$  &  73.4$_{\pm 0.3}$  &  60.8$_{\pm 0.9}$  &  42.8$_{\pm 0.5}$ & 42.8$_{\pm 1.1}$   &  28.5$_{\pm 1.0}$ &  20.2$_{\pm 0.6}$   &  14.9$_{\pm 1.3}$  &  38.9$_{\pm 1.0}$ & 23.8$_{\pm 1.1}$  & 18.3$_{\pm 1.0}$   & 14.3$_{\pm 0.3}$ & 75.5$_{\pm 0.4}$  & 70.4$_{\pm 0.3}$   \\ 
$^\dag$Rew  &   88.1$_{\pm 0.4}$  &   79.2$_{\pm 0.2}$     &  65.5$_{\pm 0.9}$  & 51.8$_{\pm 1.2}$  & 55.9$_{\pm 0.3}$  &  45.3$_{\pm 1.8}$  &  34.7$_{\pm 0.5}$  & 28.5$_{\pm 1.9}$  &  56.6$_{\pm 1.0}$   &  47.5$_{\pm 1.8}$  &  40.3$_{\pm 0.6}$  &  32.7$_{\pm 2.1}$  & 78.8$_{\pm 0.5}$   & 82.8$_{\pm 5.1}$  \\ 
$^\dag$RNF-GT &   83.0$_{\pm 3.8}$  &   73.8$_{\pm 3.3}$     &  63.8$_{\pm 1.4}$  & 48.8$_{\pm 1.7}$  & 50.7$_{\pm 0.7}$  &  36.1$_{\pm 1.0}$  &  27.6$_{\pm 1.8}$  & 21.9$_{\pm 0.5}$  &  47.5$_{\pm 0.8}$   &  32.8$_{\pm 1.0}$  &  25.4$_{\pm 0.7}$  &  19.4$_{\pm 0.6}$  & 79.0$_{\pm 1.2}$   & 74.6$_{\pm 2.9}$  \\ 
$^\dag$BiasCon  &   88.0$_{\pm 0.7}$  &   79.4$_{\pm 0.7}$     &  70.9$_{\pm 0.9}$  & 56.3$_{\pm 0.9}$  & 56.2$_{\pm 0.8}$  &  40.9$_{\pm 0.3}$  &  31.6$_{\pm 0.8}$  & 27.1$_{\pm 0.6}$  &  56.3$_{\pm 0.8}$   &  42.0$_{\pm 1.1}$  &  33.4$_{\pm 1.1}$  &  28.3$_{\pm 0.2}$  & 78.5$_{\pm 1.1}$   & 75.2$_{\pm 1.2}$    \\ 
$^\dag$GA   &  92.1$_{\pm 0.4}$ &  88.6$_{\pm 0.3}$ &  84.4$_{\pm 0.4}$ &  77.7$_{\pm 1.3}$    &  59.1$_{\pm 1.3}$   &  49.9$_{\pm 1.6}$ & 41.8$_{\pm 2.2}$  & 32.8$_{\pm 1.0}$ & 62.8$_{\pm 1.0}$  & 55.8$_{\pm 0.3}$  & 50.1$_{\pm 0.8}$ & 42.6$_{\pm 0.9}$ & 87.7$_{\pm 0.5}$  &  91.8$_{\pm 0.4}$  \\ 
\bottomrule
\end{tabular}%
}
\label{tab:last_comp}%
\end{minipage}
\end{sidewaystable*}

\subsection{Main results}
\label{sec:quan_com}

We present the main experimental results in this section. Due to the self-supervised pretext tasks will increase the training cost (but no inference latency), we split the comparison into two parts: without the self-supervised pretext tasks (in Section~\ref{sec:res_wo_ss}) and with them (in Section~\ref{sec:res_w_ss}).

\subsubsection{Without self-supervision}
\label{sec:res_wo_ss}

\textbf{The proposed method achieves better performance than others.} The overall unbiased accuracy is reported in Table~\ref{tab:overall_acc}. Vanilla models commonly fail to produce acceptable results on unbiased test sets, and the phenomenon is aggravated as $\rho$ goes larger. Different debiasing methods moderate bias propagation with varying degrees of capability. When compared to other SOTA methods, the proposed approach achieves competitive results on C-CIFAR10$^1$ and noticeable improvements on other datasets across most values of $\rho$. For instance, the vanilla model trained on C-CIFAR10$^2$ ($\rho=99\%$) only achieves 20.6\% unbiased accuracy, indicating that the model is heavily biased. While, ECS+GA leads to 50.0\% accuracy, and exceeds other prevailing debiasing methods by 3\% - 30\%. When applied to the real-world dataset B-CelebA, the proposed scheme also shows superior results, demonstrating that it can effectively deal with subtle actual biases. Though the main purpose of this work is to combat unknown biases, we find GA also achieves better performance compared to the corresponding competitors when the prior information is available. 

\begin{figure*}[htbp]
\centering
\subcaptionbox{Colored MNIST with $\rho=95\%, 98\%, 99\%, 99.5\%$ from left to right.}{
\includegraphics[width=0.16\textwidth]{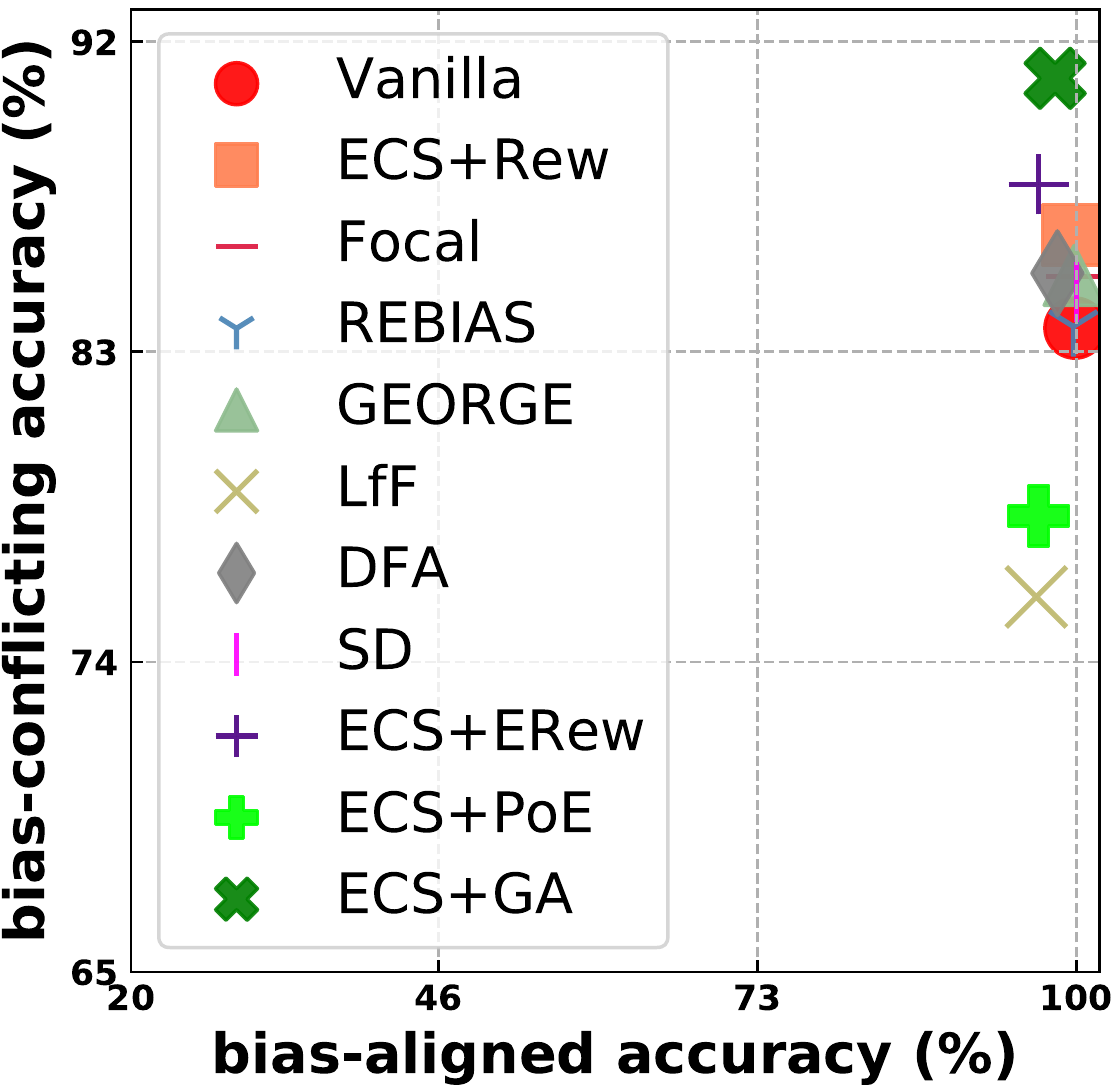} \hspace{1mm}
\includegraphics[width=0.16\textwidth]{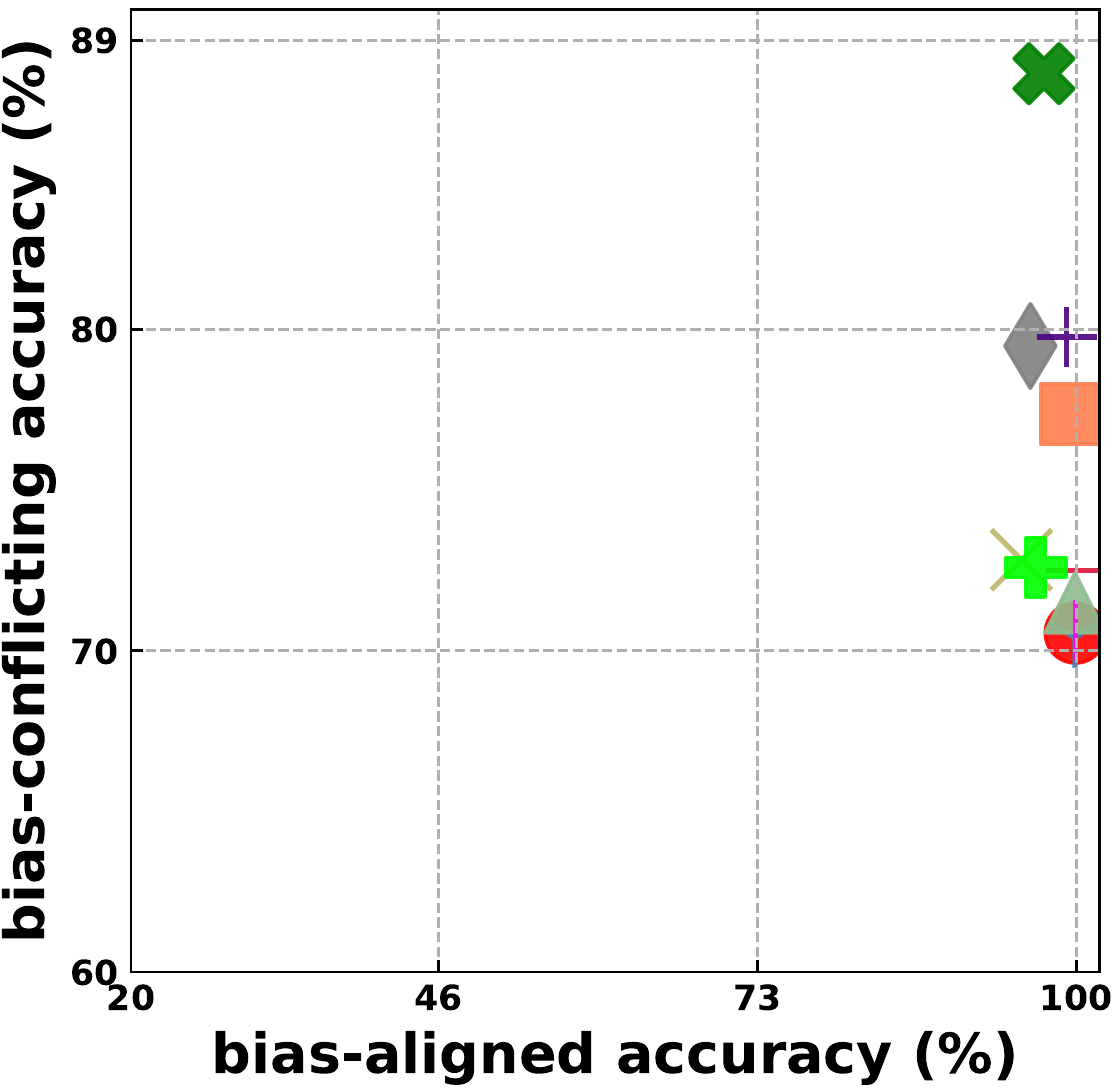} \hspace{1mm}
\includegraphics[width=0.16\textwidth]{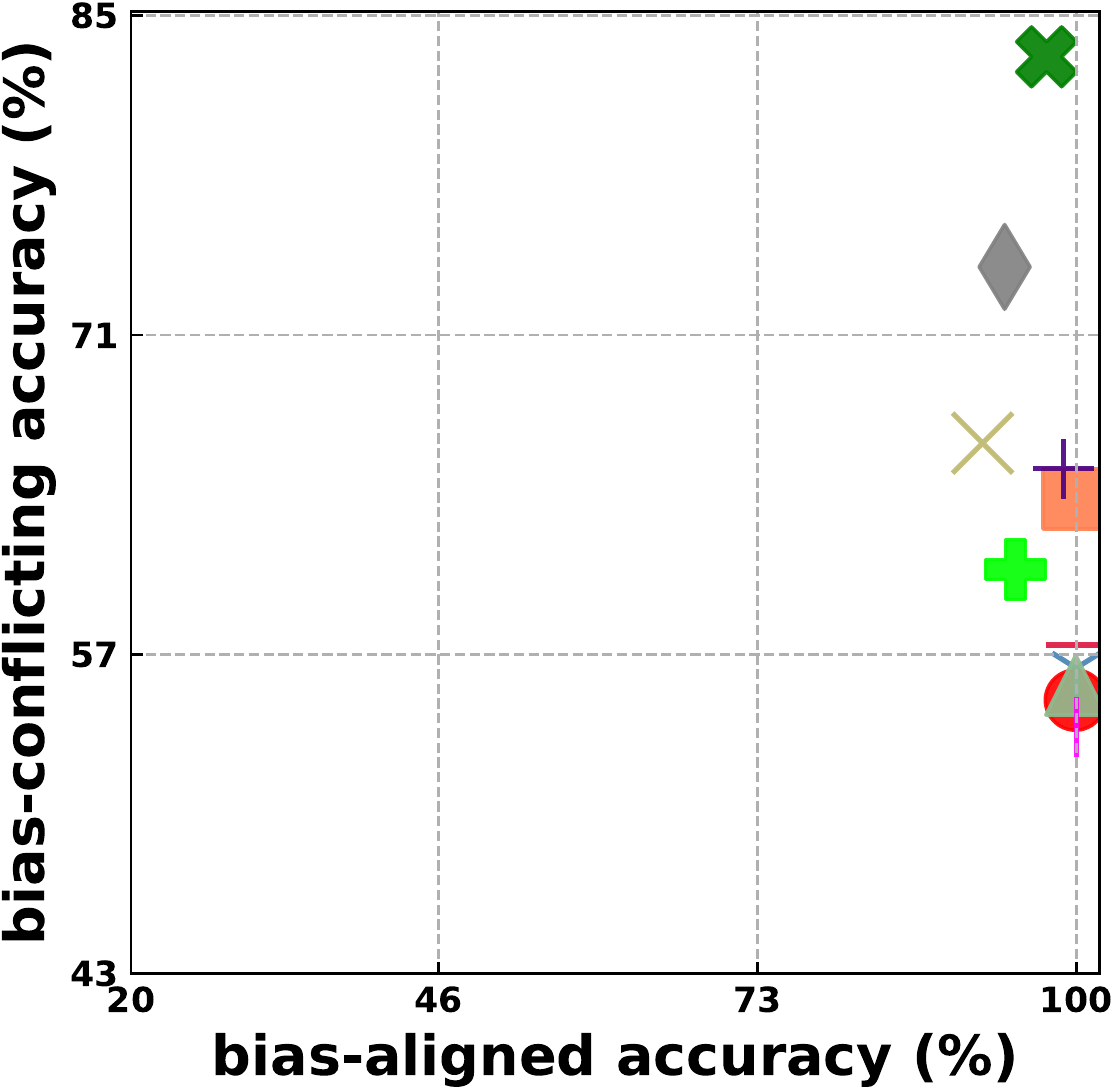} \hspace{1mm}
\includegraphics[width=0.16\textwidth]{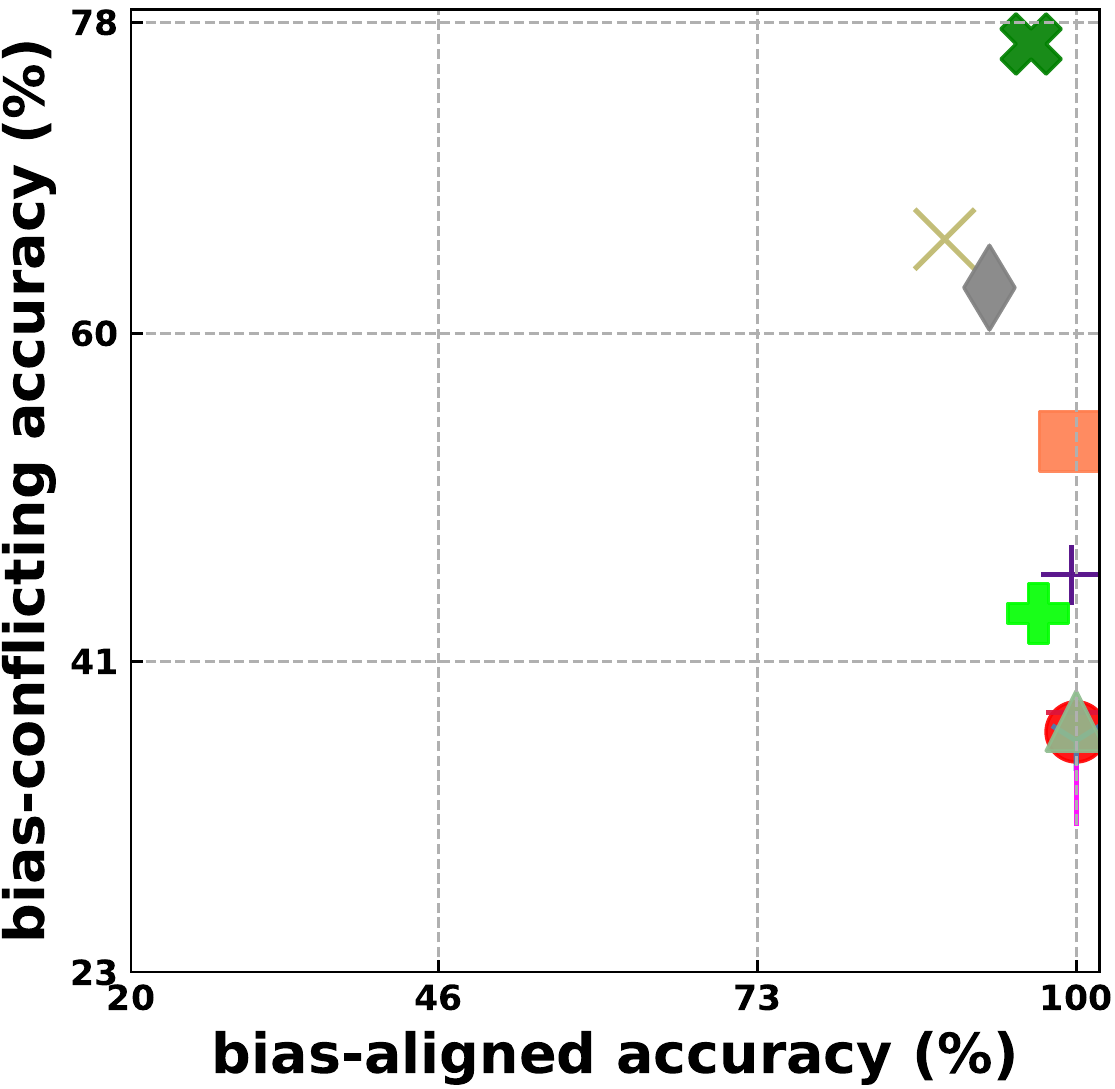} 
}
\subcaptionbox{Biased Waterbirds}{
\includegraphics[width=0.16\textwidth]{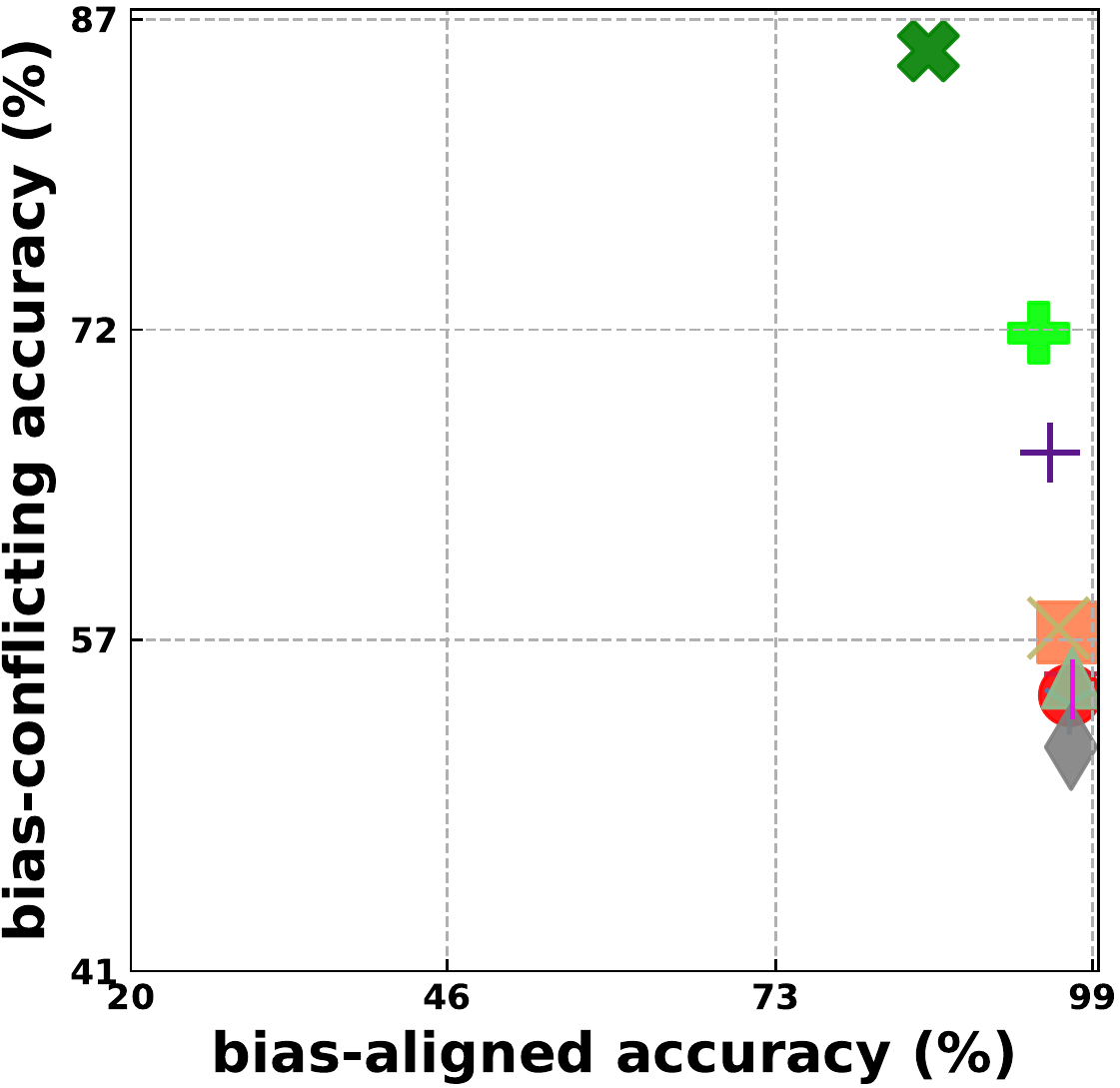}
}\\
\subcaptionbox{Corrupted CIFAR10$^1$ with $\rho=95\%, 98\%, 99\%, 99.5\%$ from left to right.}{
\includegraphics[width=0.16\textwidth]{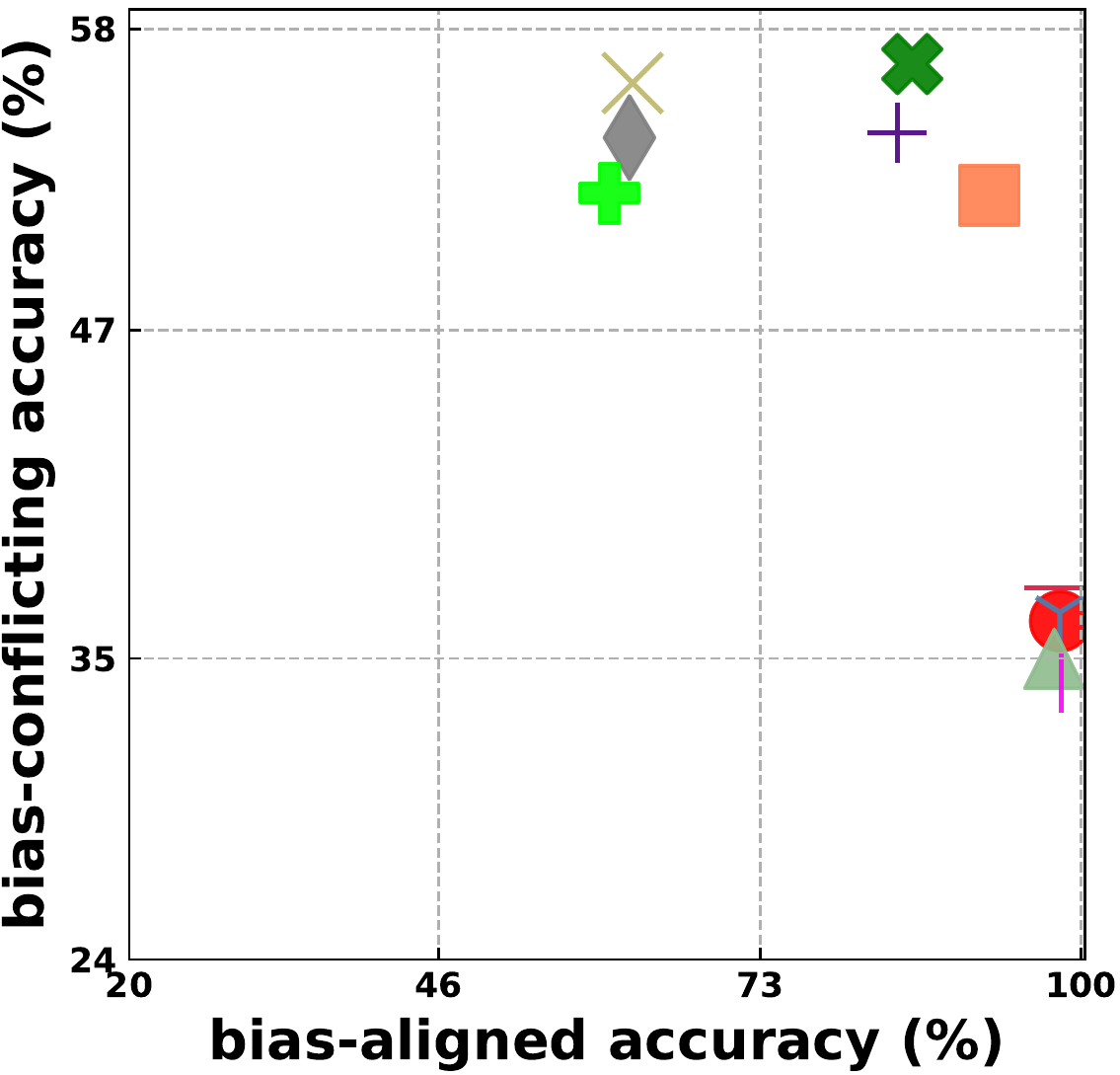}\hspace{1mm}
\includegraphics[width=0.16\textwidth]{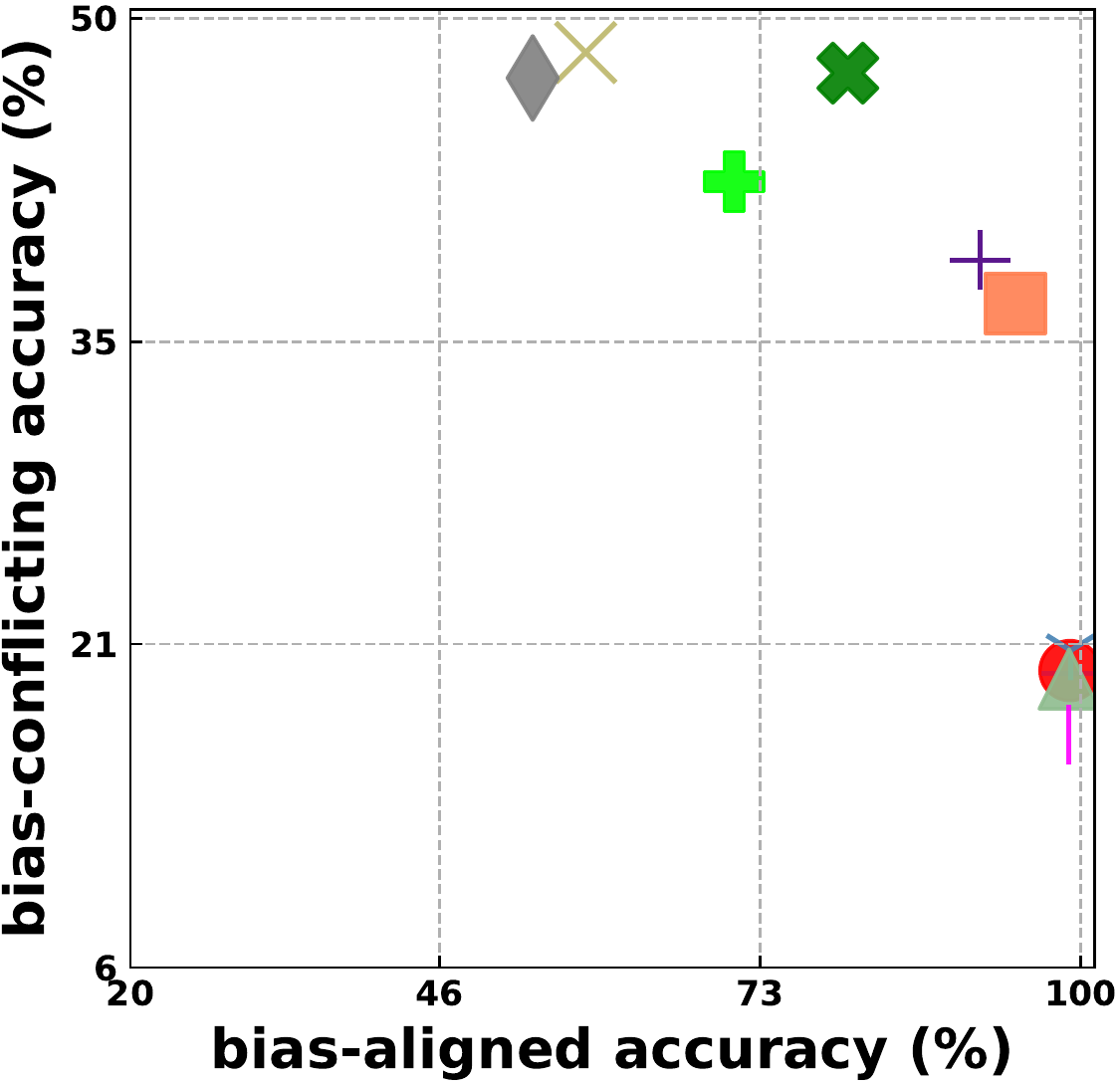}\hspace{1mm}
\includegraphics[width=0.16\textwidth]{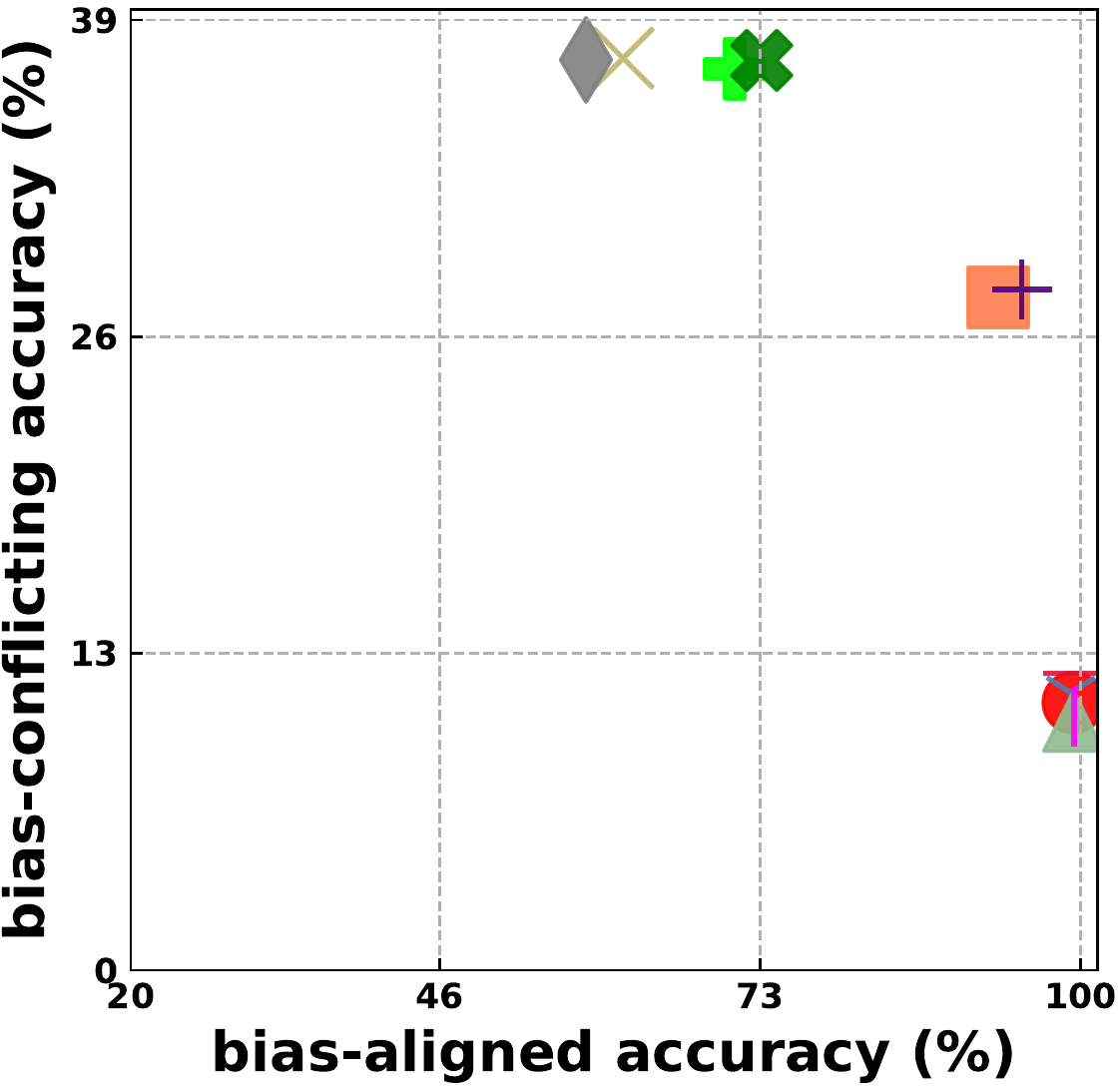}\hspace{1mm}
\includegraphics[width=0.16\textwidth]{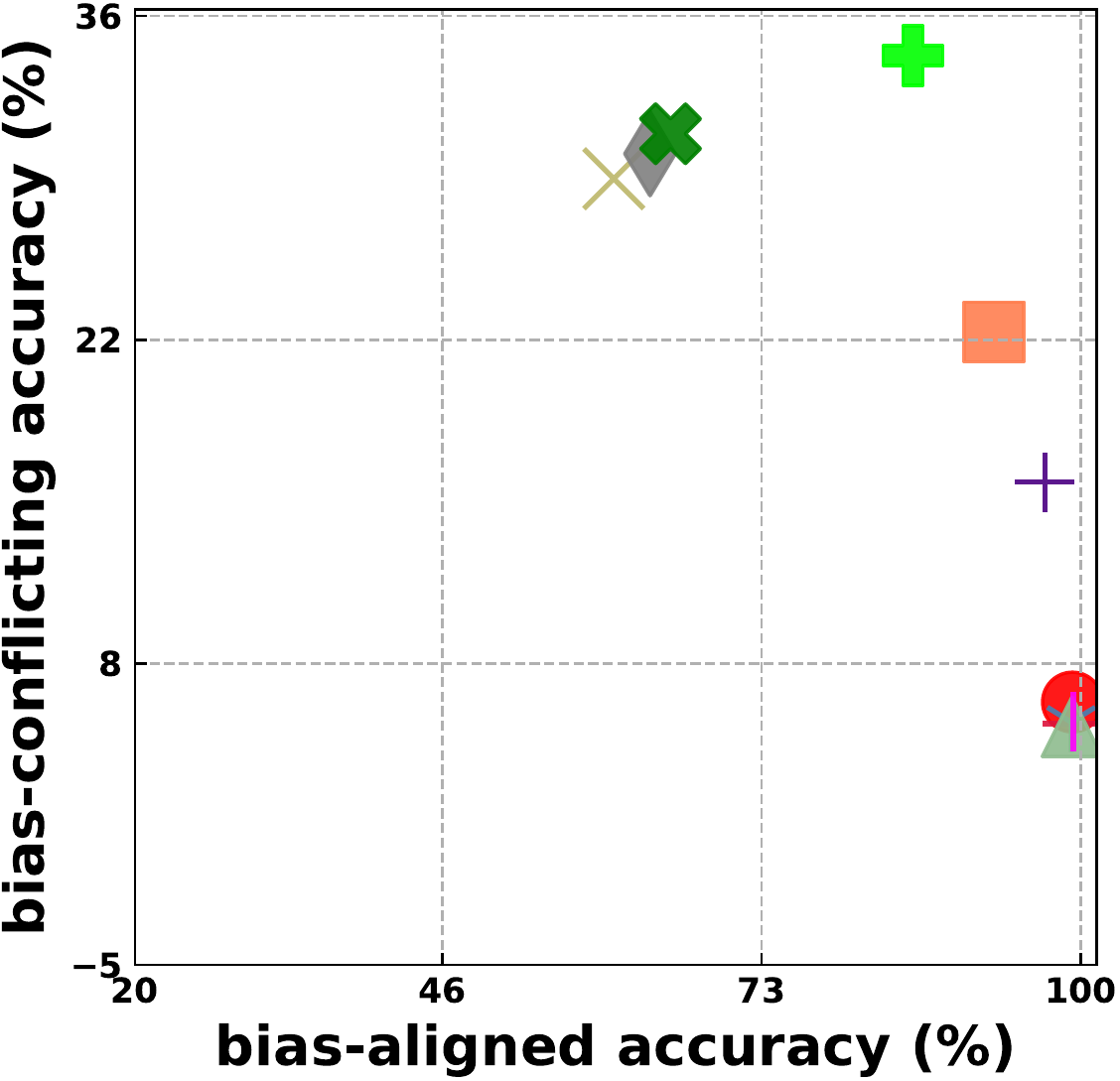}
}
\subcaptionbox{Biased CelebA}{
\includegraphics[width=0.16\textwidth]{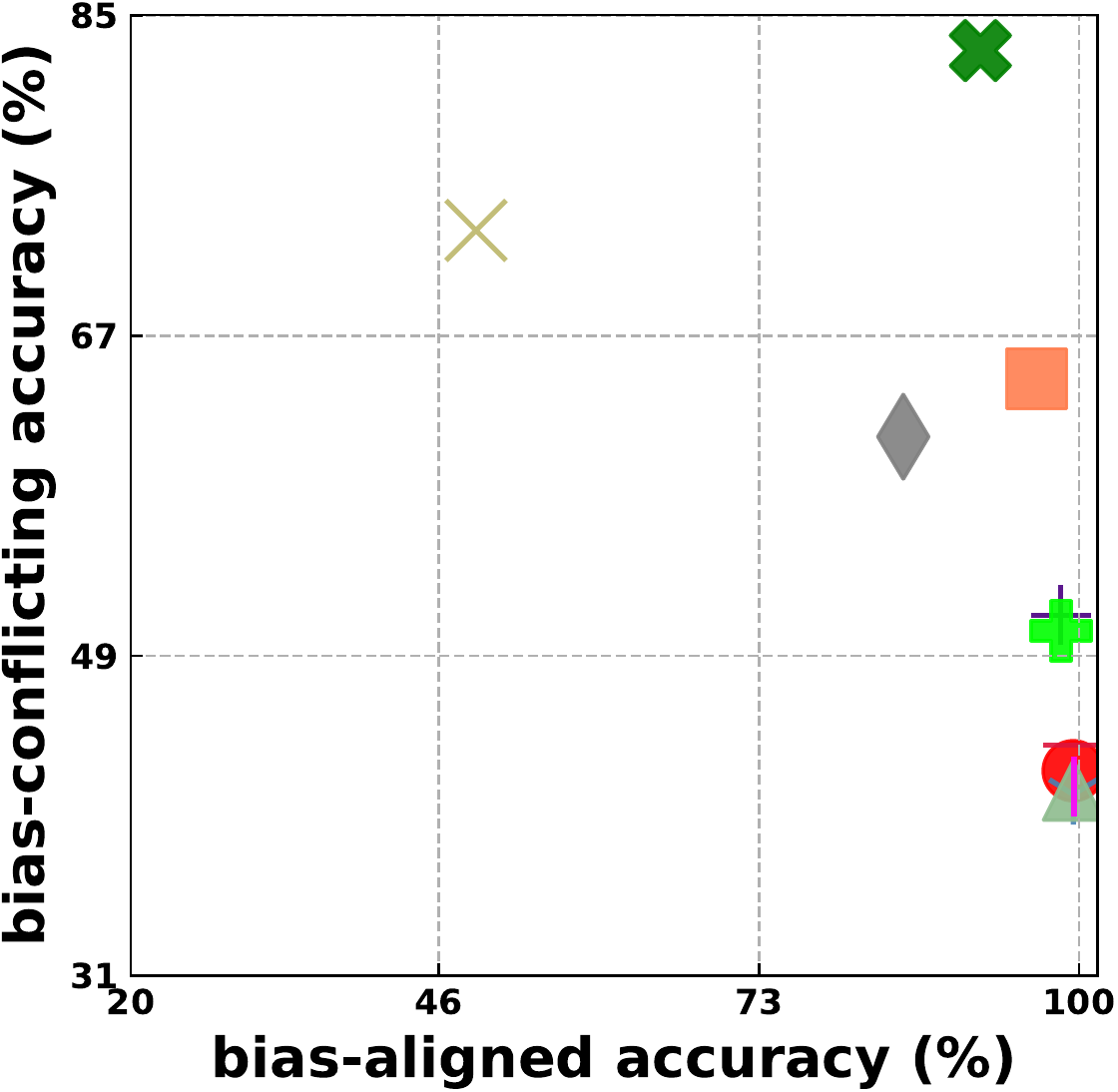}
}\\
\subcaptionbox{Corrupted CIFAR10$^2$ with $\rho=95\%, 98\%, 99\%, 99.5\%$ from left to right.}{
\includegraphics[width=0.16\textwidth]{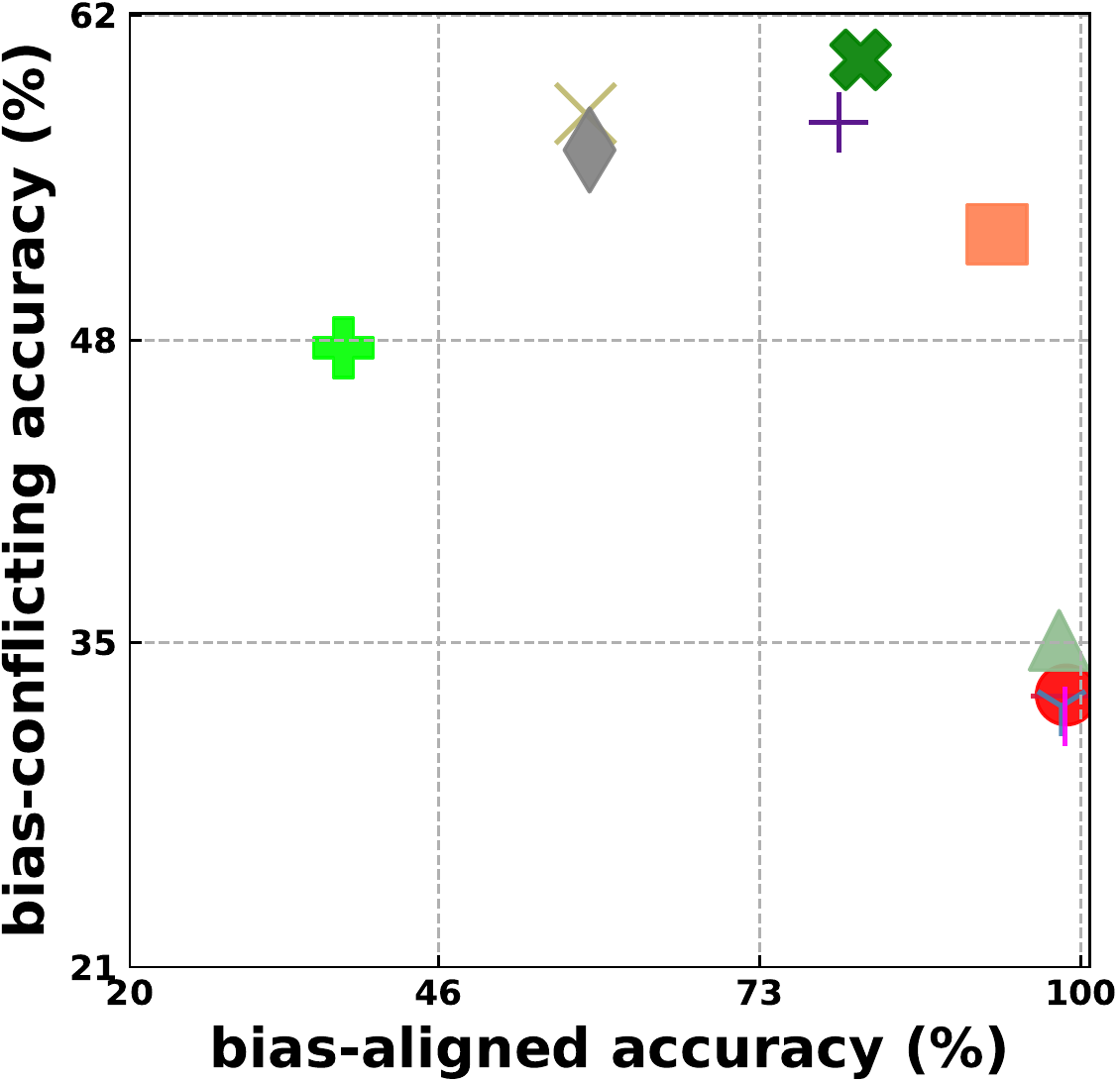}\hspace{1mm}
\includegraphics[width=0.16\textwidth]{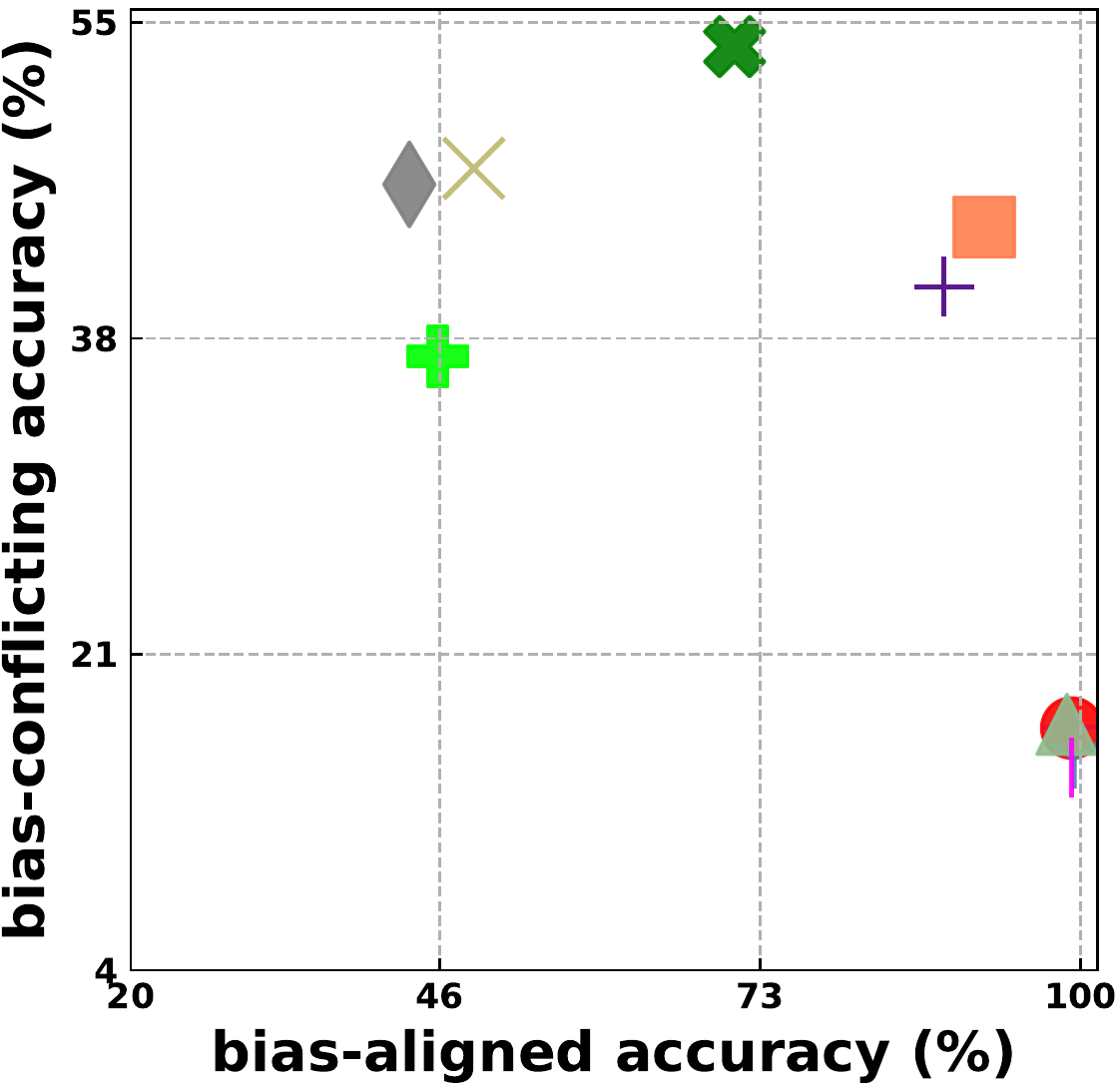}\hspace{1mm}
\includegraphics[width=0.16\textwidth]{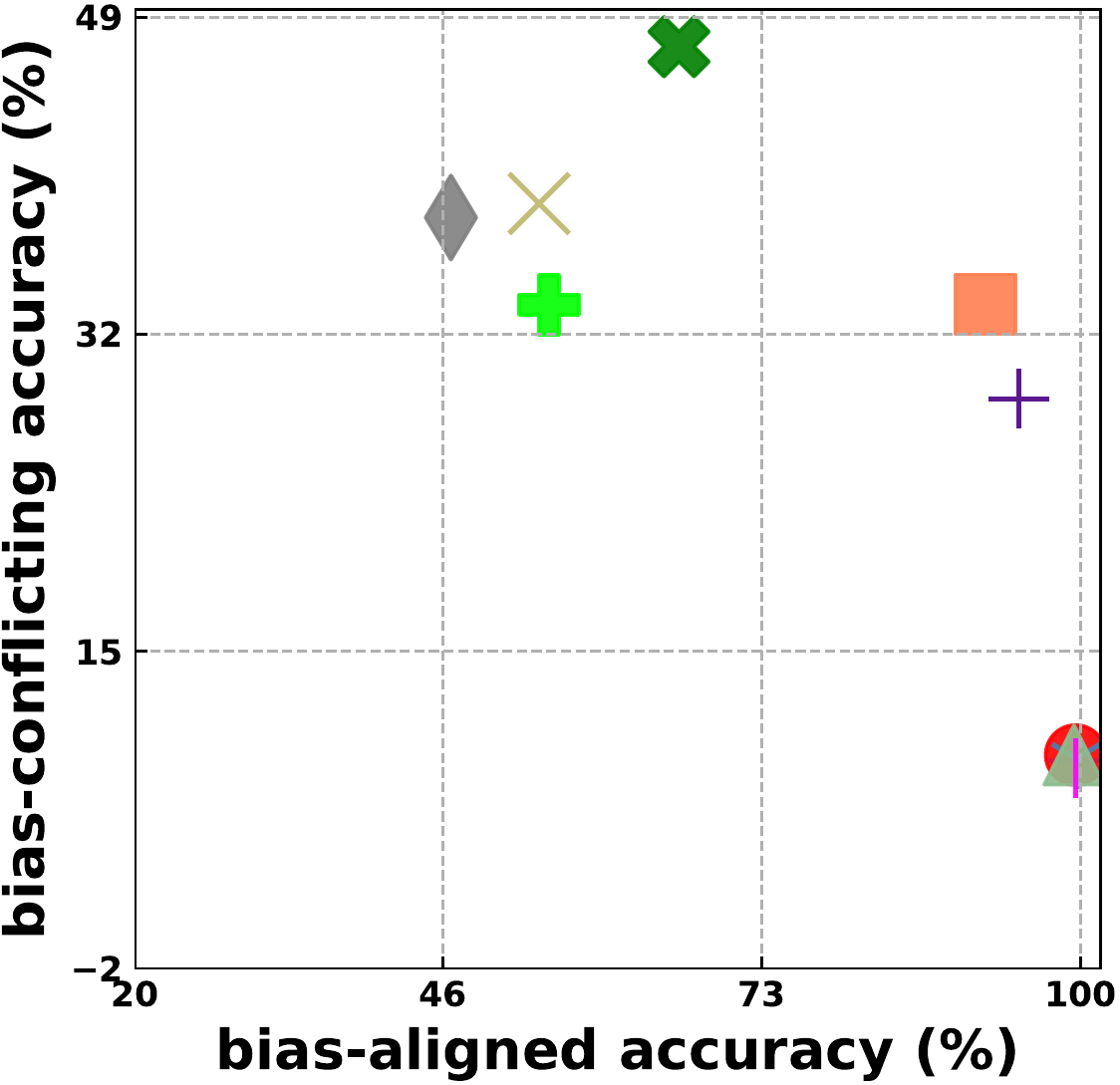}\hspace{1mm}
\includegraphics[width=0.16\textwidth]{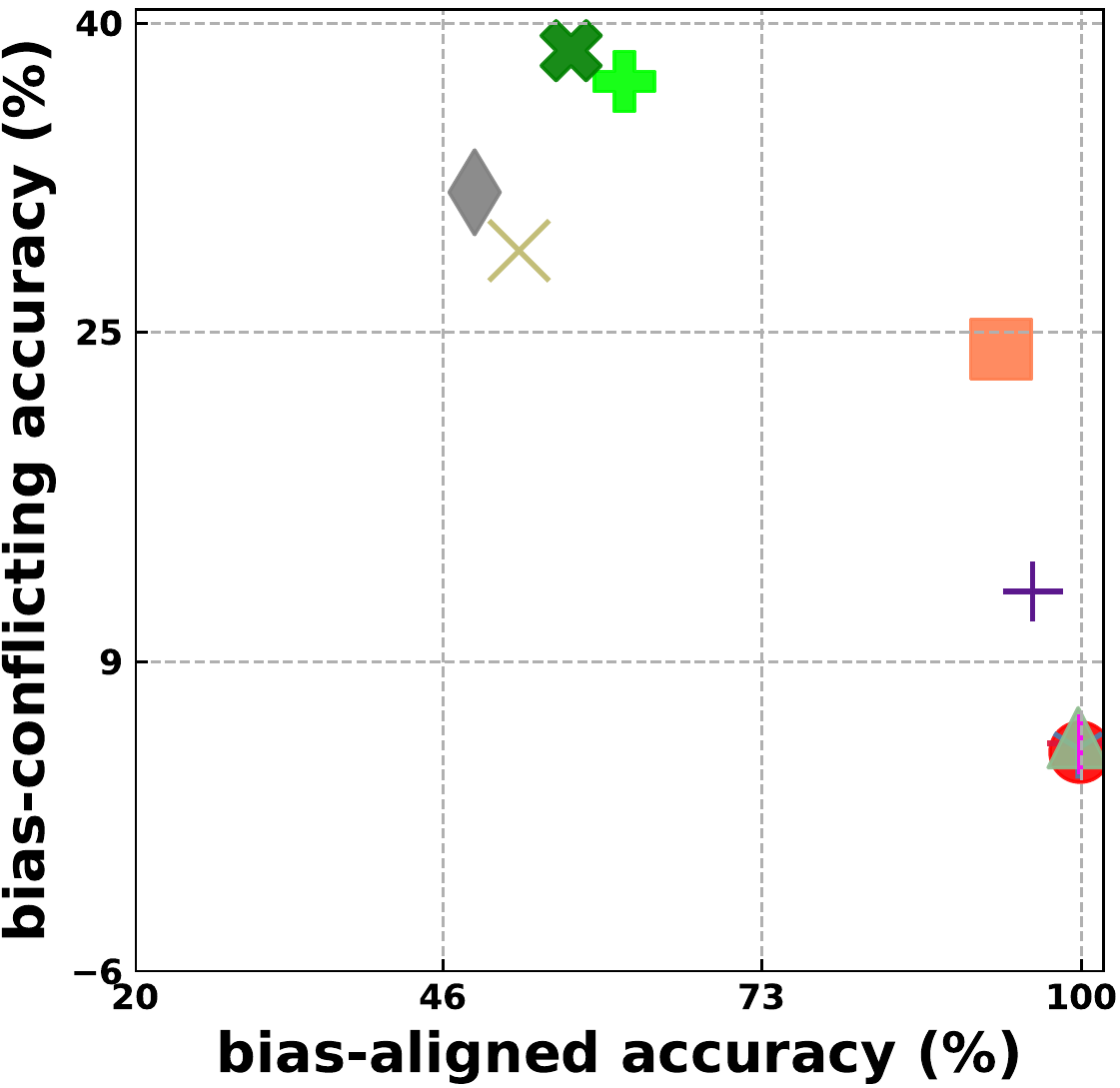}
}
\caption{Bias-aligned accuracy (horizontal-axis) and bias-conflicting accuracy (vertical-axis).}
\label{fig:acc_aligned_conflicting}
\end{figure*}

We provide the accuracy measured on the bias-aligned and bias-conflicting test samples separately in Figure~\ref{fig:acc_aligned_conflicting}. We find ECS+GA can achieve high bias-conflicting accuracy as well as bias-aligned accuracy mostly, leading to superior overall unbiased performance. Note that, too high bias-aligned accuracy is not always good. Though the vanilla model can obtain a very high illusory bias-aligned accuracy assisted with biases, it does not learn intrinsic features as shown in Figure~\ref{fig:cam_supp1}, leading to extremely pool out-of-distribution generalization. As an instance, although the vanilla model trained on Corrupted CIFAR10$^2$ ($\rho=99\%$) achieves high bias-aligned accuracy (99.6\%), the value (99.6\%) actually reflects the model's ability to discriminate bias attribute rather than target attribute. In fact, when training on an unbiased training set ($\rho=10\%$), the corresponding accuracy is only 79.5\%, reflecting that the real target task is harder to learn than the spurious one.

\textbf{Plain reweighting is an important baseline.} We find Rew (and ECS+Rew) can achieve surprising results compared with recent state-of-the-art methods, while it is overlooked by some studies. The results also indicate that explicitly balancing the bias-aligned and bias-conflicting samples is extremely important and effective.

\begin{table}[t]
\centering
\caption{Performance in terms of DP and EqOdd on Biased Waterbirds (left) and Biased CelebA (right).}
\setlength\tabcolsep{3pt}
\resizebox{0.999\columnwidth}{!}{
\begin{tabular}{l|cc|cc}
\toprule
& DP $\uparrow$   & EqOdd $\uparrow$ & DP $\uparrow$   & EqOdd $\uparrow$\\
\midrule
Vanilla & 0.57$_{\pm 0.01}$ & 0.57$_{\pm 0.01}$ & 0.43$_{\pm 0.01}$ & 0.43$_{\pm 0.02}$ \\
LfF   & 0.63$_{\pm 0.03}$ & 0.61$_{\pm 0.03}$ & \textbf{0.80}$_{\pm 0.06}$ & 0.76$_{\pm 0.07}$ \\
DFA   & 0.55$_{\pm 0.01}$ & 0.55$_{\pm 0.02}$ & 0.69$_{\pm 0.01}$ & 0.76$_{\pm 0.06}$ \\
ECS+Rew & 0.61$_{\pm 0.02}$ & 0.60$_{\pm 0.01}$ & 0.59$_{\pm 0.01}$ & 0.68$_{\pm 0.18}$ \\
ECS+GA & \textbf{0.99}$_{\pm 0.01}$ & \textbf{0.99}$_{\pm 0.01}$ & 0.73$_{\pm 0.02}$ & \textbf{0.91}$_{\pm 0.02}$ \\
\bottomrule
\end{tabular}%
}
\label{tab:fairness}%
\end{table}

\begin{table*}[t]
\centering
\caption{The effectiveness of self-supervision. The overall unbiased accuracy and standard deviation of the last epoch over 3 runs (\%) are reported. Best results with unknown biases are in bold. $^\dag$ indicates that they require prior knowledge regarding biases.}
\resizebox{0.97\textwidth}{!}{
\begin{tabular}{l|cccc|cccc|c}
\toprule
& \multicolumn{4}{c|}{Corrupted CIFAR10$^1$} & \multicolumn{4}{c|}{Corrupted CIFAR10$^2$} & B-Birds \\
\multicolumn{1}{c|}{$\rho$} & 95\% & 98\% & 99\% & 99.5\% & 95\% & 98\% & 99\% & 99.5\% & 95\% \\
\midrule
Vanilla & 42.6$_{\pm0.4}$ & 27.7$_{\pm1.0}$ & 19.8$_{\pm1.0}$ & 15.6$_{\pm0.8}$ & 39.3$_{\pm0.6}$ & 25.3$_{\pm1.3}$ & 18.5$_{\pm0.5}$ & 14.2$_{\pm0.3}$ & 75.7$_{\pm0.8}$ \\
\quad+SS & 51.8$_{\pm0.8}$ & 38.5$_{\pm1.2}$ & 29.8$_{\pm0.9}$ & 23.7$_{\pm0.5}$ & 47.9$_{\pm1.4}$ & 34.5$_{\pm1.0}$ & 26.4$_{\pm0.5}$ & 20.2$_{\pm0.2}$ & 80.8$_{\pm0.6}$ \\
ECS+GA & 59.6$_{\pm0.5}$ & 50.8$_{\pm1.0}$ & 40.9$_{\pm0.3}$ & 34.3$_{\pm0.4}$ & 62.2$_{\pm0.6}$ & 55.4$_{\pm2.6}$ & 49.3$_{\pm1.2}$ & 40.5$_{\pm0.5}$ & 85.5$_{\pm0.9}$ \\
\quad+SS & \textbf{63.5}$_{\pm0.6}$ & \textbf{55.7}$_{\pm2.2}$ & \textbf{50.1}$_{\pm0.6}$ & \textbf{44.5}$_{\pm0.4}$ & \textbf{65.1}$_{\pm1.9}$ & \textbf{59.0}$_{\pm1.2}$ & \textbf{55.0}$_{\pm1.1}$ & \textbf{48.6}$_{\pm0.8}$ & \textbf{86.9}$_{\pm0.3}$ \\
\hline \hline
$^\dag$GA & 59.1$_{\pm1.3}$ & 49.9$_{\pm1.6}$ & 41.8$_{\pm2.2}$ & 32.8$_{\pm1.0}$ & 62.8$_{\pm1.0}$ & 55.8$_{\pm0.3}$ & 50.1$_{\pm0.8}$ & 42.6$_{\pm0.9}$ & 87.7$_{\pm0.5}$ \\
\quad+SS & 63.6$_{\pm0.6}$ & 57.0$_{\pm0.5}$ & 52.2$_{\pm1.2}$ & 45.6$_{\pm1.1}$ & 64.7$_{\pm0.9}$ & 59.9$_{\pm0.5}$ & 56.5$_{\pm0.4}$ & 50.6$_{\pm0.7}$ & 89.5$_{\pm0.0}$ \\
\bottomrule
\end{tabular}%
}
\label{tab:res_ss}%
\end{table*}%

\textbf{Early-stopping is not necessary for GA to select models.} Plain reweighting requires strong regularizations such as early-stopping to produce satisfactory results~\citep{byrd2019effect,Sagawa*2020Distributionally}, implying that the results are not stable. Due to the nature of combating unknown biases, the unbiased validation set is not available, thus recent studies choose to report the best results among epochs~\citep{nam2020learning,kim2021learning} for convenient comparison. We follow this evaluation protocol in Table~\ref{tab:overall_acc}. However, in the absence of prior knowledge, deciding when to stop can be troublesome, thus some results in Table~\ref{tab:overall_acc} are excessively optimistic. We claim that if the network can attain dynamic balance throughout the training phase, such early-stopping may not be necessary. We further provide the last epoch results in Table~\ref{tab:last_comp} to validate it. We find that some methods suffer from serious performance degradation. On the contrary, GA achieves steady results (with the same and fair training configurations). In other words, our method shows superiority under two model selection strategies simultaneously.

\textbf{The proposed method has strong performance on fairness metrics as well.} As shown in Table~\ref{tab:fairness}, the proposed method also obtains significant improvement in terms of DP and EqOdd. These results further demonstrate that the proposed method is capable of balancing bias-aligned and bias-conflicting samples, as well as producing superior and impartial results.

\subsubsection{With self-supervision}
\label{sec:res_w_ss}

\textbf{Self-supervision improves vanilla training.} As shown in Table~\ref{tab:res_ss}, the self-supervised pretext tasks achieve obvious improvement over vanilla training, demonstrating the effectiveness of self-supervision in the context of debiasing.

\textbf{Self-supervision also promotes advanced debiasing methods.} As shown in Table~\ref{tab:res_ss}, the self-supervised pretext tasks also lead to significant gains on the basis of different debiasing methods and on a variety of datasets. When the training is heavily biased, the improvement is very significant, \textit{e.g.}, 10.2\% and 8.1\% gains on C-CIFAR10$^1$ and C-CIFAR10$^2$ ($\rho = 99.5\%$) beyond our method ECS+GA, respectively. Due to the low diversity of the bias-conflicting samples within the severely biased training data, the gain of ECS+GA may be limited, but self-supervision helps the model discover more general characteristics from the adequate bias-aligned examples.

\section{Further analysis}
\label{sec:further_ana}

\subsection{ECS shows superior ability to mine bias-conflicting samples}

\begin{figure}[t]
\centering
\includegraphics[width=0.44\columnwidth]{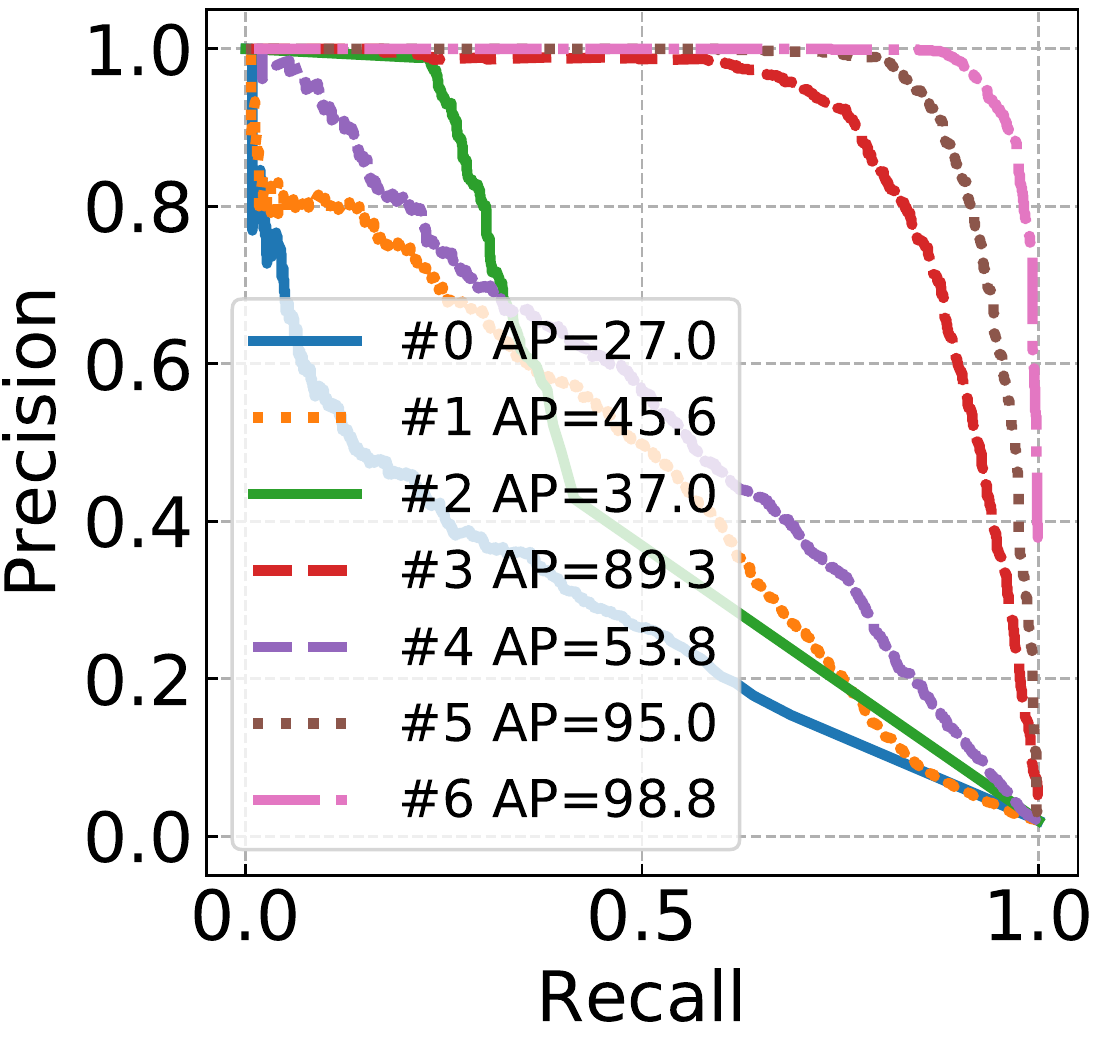} \hspace{2mm}
\includegraphics[width=0.44\columnwidth]{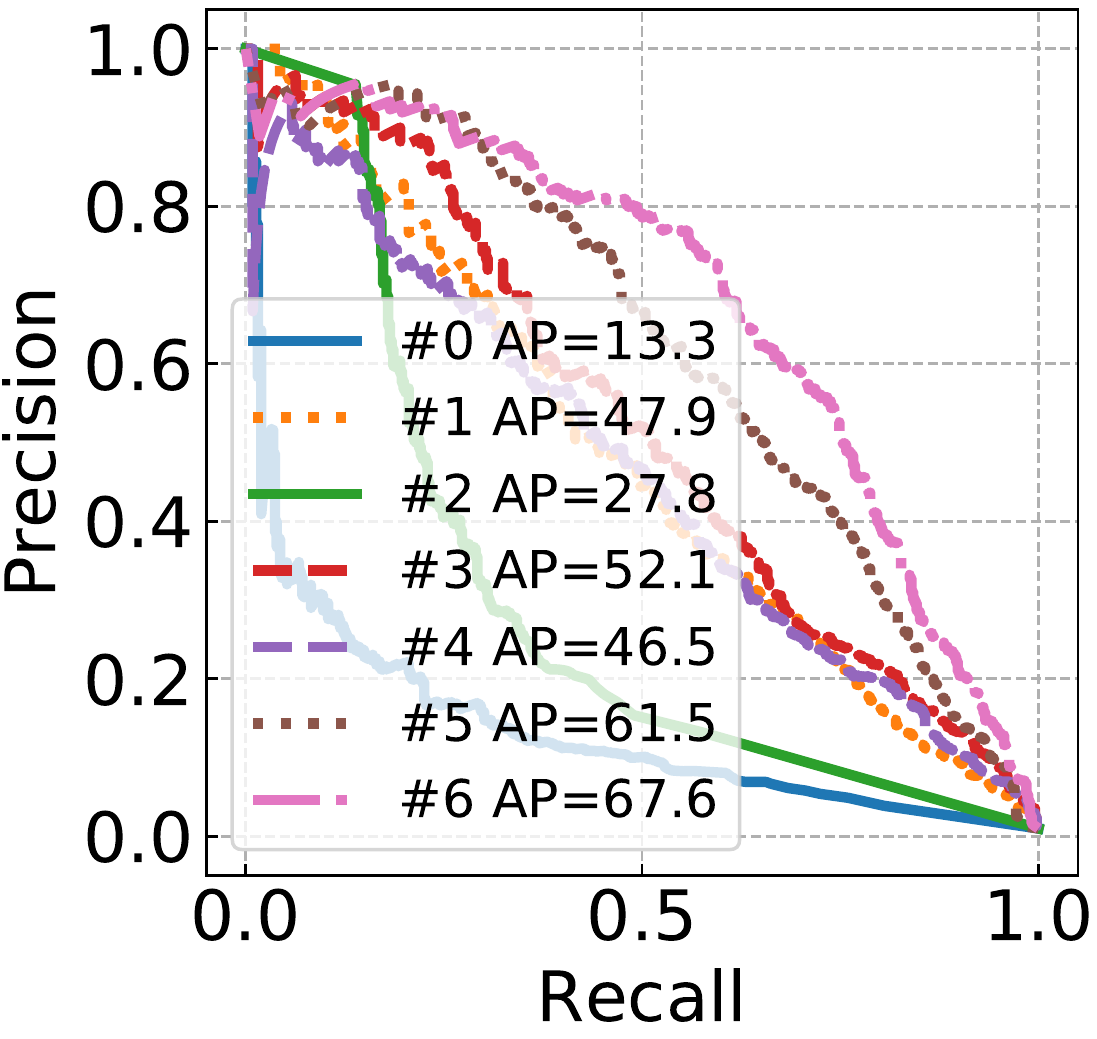}
\caption{Precision-recall curves of different bias-conflicting scoring methods on Colored MNIST ($\rho=98\%$, left) and Biased CelebA (right).}
\label{fig:pr_curves}
\end{figure}

We separately verify the effectiveness of each component of ECS on C-MNIST ($\rho=98\%$) and B-CelebA. A good bias-conflicting scoring method should prompt superior precision-recall curves for the mined bias-conflicting samples, \textit{i.e.}, give real bias-conflicting (aligned) samples high (low) scores. Therefore, we provide the average precision (AP) in Table~\ref{tab:variants} (P-R curves are illustrated in Figure~\ref{fig:pr_curves}). When comparing \#0, \#4, \#5, and \#6, we observe that epoch-ensemble, confident-picking and peer model all can improve the scoring method. In addition, as shown in Table~\ref{tab:overall_acc}, ECS+GA achieves results similar to GA with the help of ECS; ERew, PoE, and Rew combined with ECS also successfully alleviate biases to some extent, demonstrating that the proposed ECS is feasible, robust, and can be adopted in a variety of debiasing approaches.

\begin{table*}[t]
\centering 
\captionof{table}{Average precision (\%) of the mined bias-conflicting samples. VM: scoring with vanilla model.}
\resizebox{0.85\textwidth}{!}{
\begin{tabular}{l|l|c c c|c c}
\toprule
&  & Epoch-Ensemble & Confident-Picking & Peer Model  & C-MNIST & B-CelebA \\
\midrule
\#0   & VM &  &   & & 27.0  & 13.3  \\
\#1   & ES (in JTT) &  &   &   & 45.6  & 47.9  \\
\#2   & GCE (in LfF) &   &  &   & 37.0  & 27.8  \\
\#3   & GCE + EE & \checkmark &  &   & 89.3  & 52.1  \\
\hline \hline
\#4   & ECS (Ours) & \checkmark &   &   & 53.8  & 46.5  \\
\#5   & ECS (Ours)  & \checkmark & \checkmark &  & 95.0  & 61.5  \\
\#6   & ECS (Ours)   & \checkmark & \checkmark & \checkmark & \textbf{98.8}  & \textbf{67.6}  \\
\bottomrule
\end{tabular}%
}
\label{tab:variants}%
\end{table*}

\begin{figure}[t]
\centering
\includegraphics[width=0.46\columnwidth]{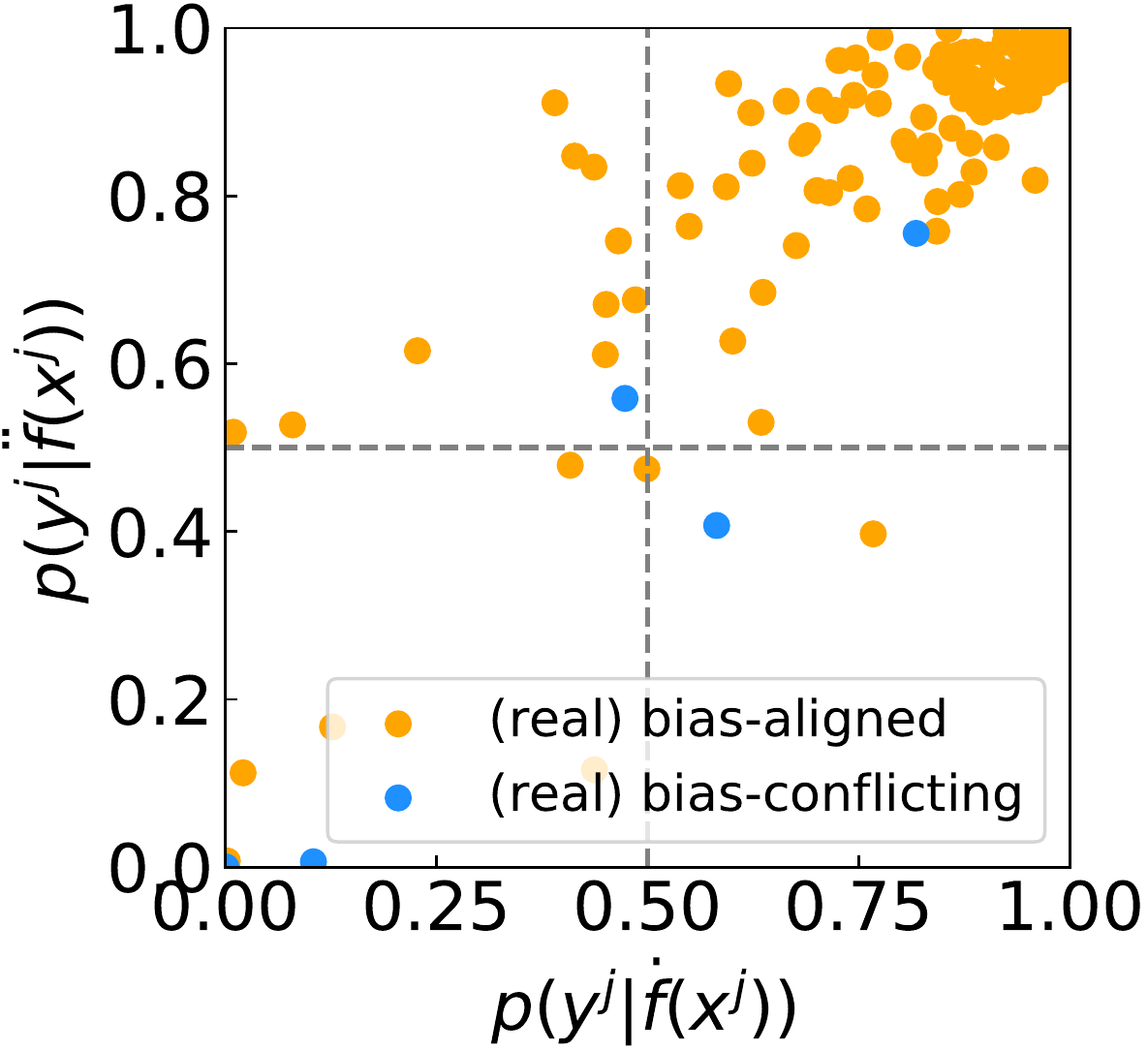} \hspace{2mm}
\includegraphics[width=0.46\columnwidth]{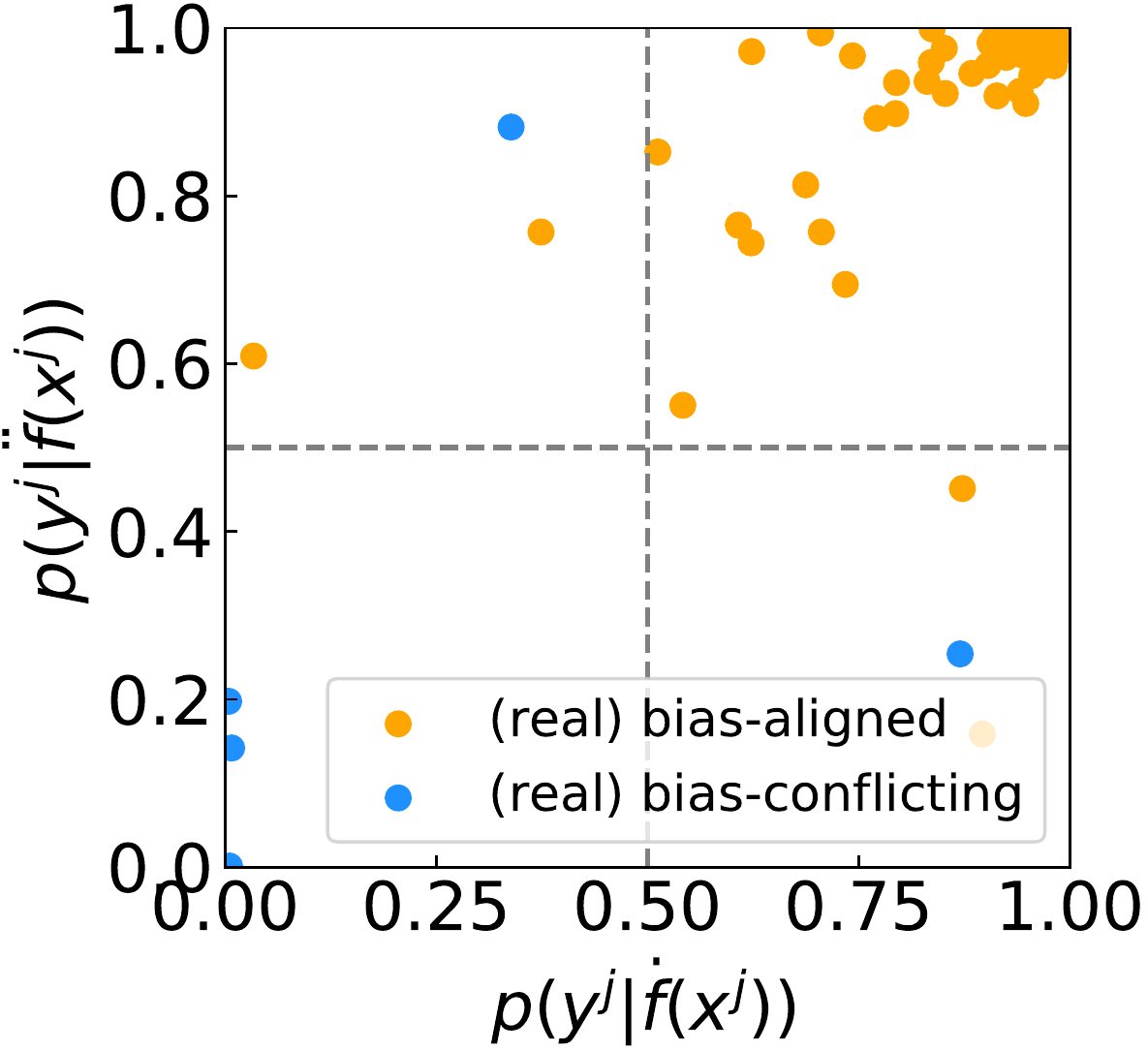}
\caption{Visualizations of the predictions of peer models (during training).}
\label{fig:peer_vis}
\end{figure}

We further compare the methods: \#1 collecting results with early-stopping (ES) in JTT~\citep{liu2021just}, \#2 training auxiliary biased model with GCE loss in LfF (and \#3 collecting results with epoch-ensemble on top of it). When comparing \#1 and \#4, both early-stopping and epoch-ensemble can reduce the overfitting to bias-conflicting samples when training biased models, yielding more accurate scoring results. However, early-stopping is laborious to tune~\citep{liu2021just}, whereas epoch-ensemble is more straightforward and robust. From \#2 and \#3, we see that epoch-ensemble can also enhance other strategies. Comparing \#3 and \#5, GCE loss is helpful, while confident-picking gains better results. Noting that though co-training with peer model raises some costs, it is not computationally complex and can yield significant benefits (\#6), and even without peer model, \#5 still outperform previous ways. Peer models are expected to better prevent bias-conflicting samples from affecting the training, so we can get better auxiliary biased models. Though the only difference between peer models is initialization in our experiments, as DNNs are highly nonconvex, different initializations can lead to different local optimal~\citep{han2018co}. We provide the visualizations of the predictions of peer models (during training) in Figure~\ref{fig:peer_vis}.

\begin{table}[t]
\centering
\caption{Results (precision, recall, and unbiased accuracy) of GA combined with different bias-conflicting scoring methods.}
\setlength\tabcolsep{3pt}
\resizebox{0.49\textwidth}{!}{
\begin{tabular}{ c | c  c  c |   c  c  c}
\toprule
& \multicolumn{3}{ c|}{Colored MNIST ($\rho=98\%$)}    & \multicolumn{3}{ c}{Biased CelebA} \\
&  P (\%)  &    R (\%)   &      Acc (\%)   &    P (\%)    &    R (\%)  & Acc (\%)  \\
\midrule
\#0+GA   & 90.9  & 0.8  & 49.2   & 77.8  & 1.5  & 78.9 \\
\#1+GA      & 96.4  & 2.2 & 58.9 &  79.2  & 20.6  & 86.2 \\
\#2+GA    &  98.6  &  23.4  & 76.6     & 79.6 &  16.9 & 84.2 \\
\#3+GA    & 98.9  & 45.7  & 86.3    & 79.3  & 27.3  & 86.2 \\
\#4+GA     & 98.5  & 5.5  & 62.7     & 79.8  & 16.2  & 82.8 \\
\#5+GA     & 99.8  & 67.9  & 88.9      & 79.2  & 39.6  & 87.9 \\
\#6+GA   & 99.9  & \textbf{84.8}  & \textbf{89.5}  & 79.1 & \textbf{50.0}  & \textbf{89.1} \\
\hline \hline
Vanilla   & -  & -  & 73.6    & -  & -  & 77.4 \\
\bottomrule
\end{tabular}%
}
\label{tab:bga_diff_mining}%
\end{table}%

We also provide the results of GA combined with the above bias-conflicting scoring variants in Table~\ref{tab:bga_diff_mining} (for fairness, all methods are compared under a similar precision), which show all the proposed components contribute to a more robust model in stage \uppercase\expandafter{\romannumeral2}. Finally, we provide the precision and recall of our mined bias-conflicting samples with the help of ECS and the typical value of $\tau$ (0.8) in Table~\ref{tab:complete_pr}.

\begin{table}[t]
\centering
\caption{Precision and recall (\%) of the mined bias-conflicting samples with ECS.}
\begin{tabular}{llcc}
\toprule
&   $\rho$    & Precision & Recall \\
\midrule
& 95\% & 99.5  & 70.0  \\
Colored  & 98\% & 99.9  & 84.8  \\
MNIST & 99\% & 99.8  & 89.3  \\
& 99.5\% & 99.6  & 92.7  \\
\midrule
& 95\% & 96.3  & 92.4  \\
Corrupted  & 98\% & 92.3  & 94.2  \\
CIFAR10$^1$ & 99\% & 87.6  & 93.2  \\
& 99.5\% & 76.0  & 94.8  \\
\midrule
& 95\% & 99.2  & 94.2  \\
Corrupted & 98\% & 98.3  & 94.9  \\
CIFAR10$^2$ & 99\% & 97.3  & 95.0  \\
& 99.5\% & 93.3  & 94.8  \\
\midrule
B-Birds & 95\% &  77.7  &  65.2 \\
\midrule
B-CelebA & 99\% & 79.1 &  50.0 \\
\bottomrule
\end{tabular}%
\label{tab:complete_pr}%
\end{table}%

\begin{figure*}[t]
\centering
\includegraphics[width=0.5\columnwidth]{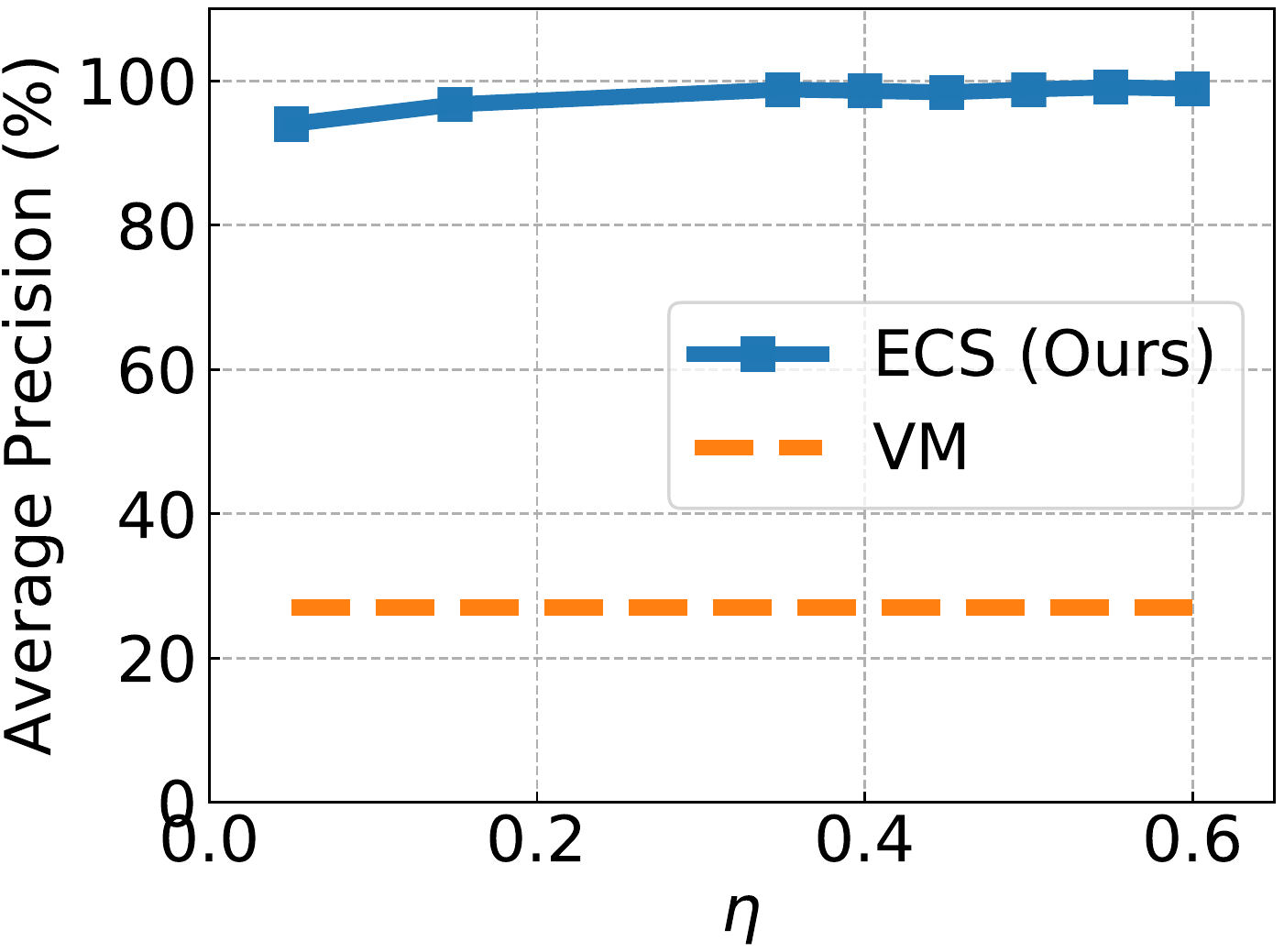} \hspace{4mm}
\includegraphics[width=0.5\columnwidth]{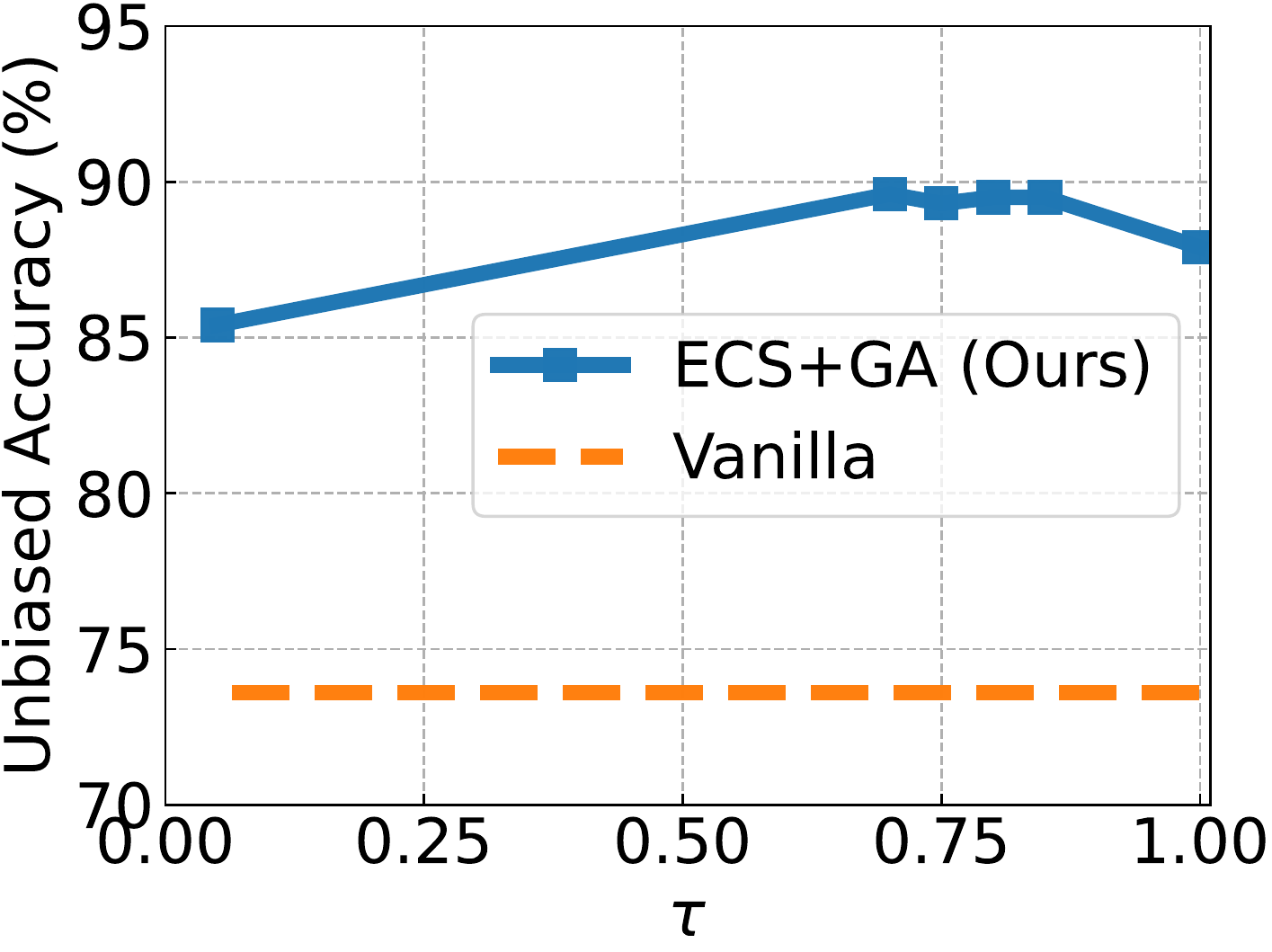} \hspace{4mm}
\includegraphics[width=0.5\columnwidth]{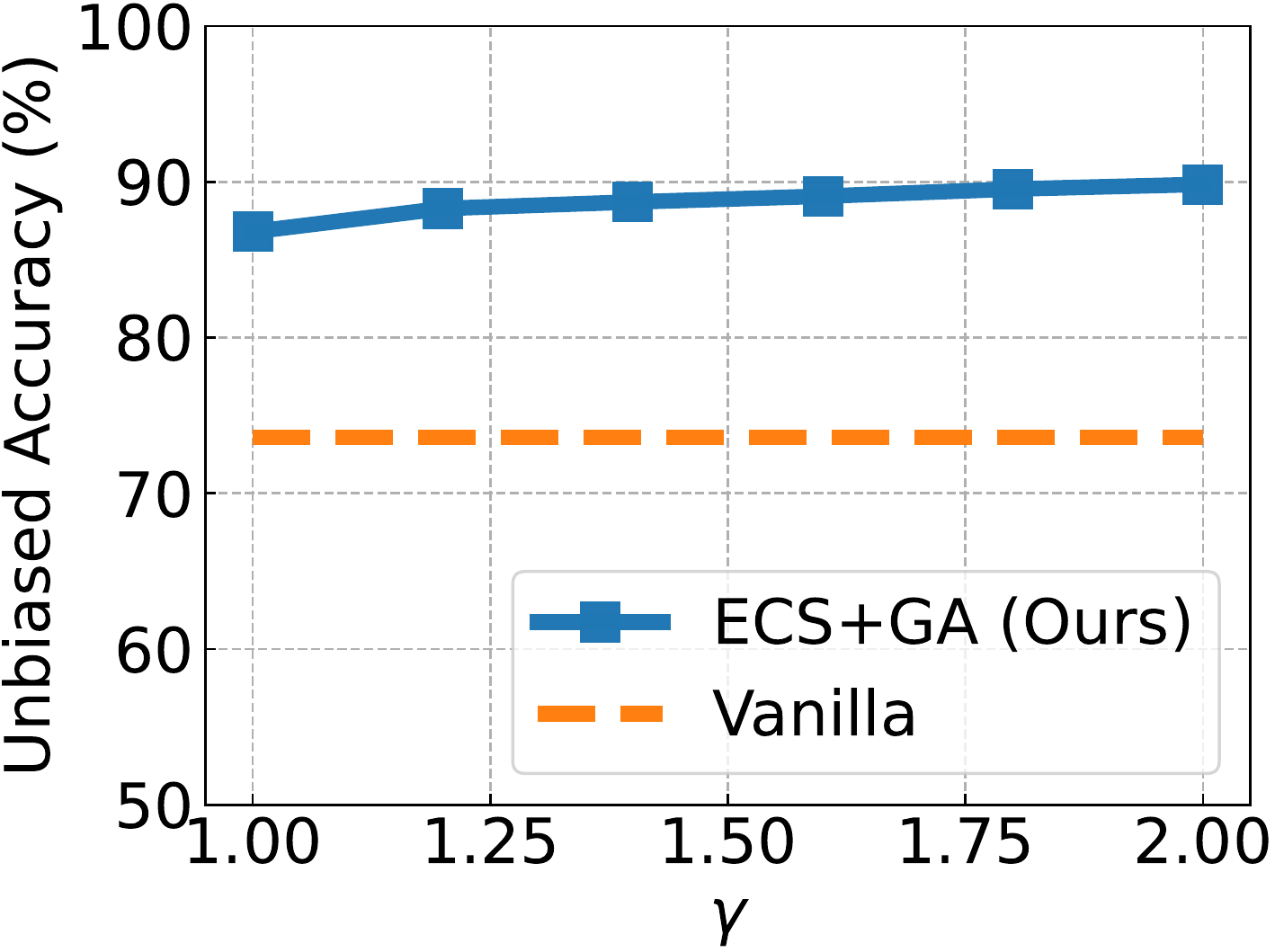}
\caption{Ablation on thresholds $\eta$, $\tau$ and balance ratio $\gamma$.}
\label{fig:vary_param}
\end{figure*}

\subsection{Differences between GA and its counterparts}

Focal loss, LfF, DFA, and ERew just reweight a sample with the information from itself (individual information), different from them, GA, as well as Rew, use global information within one batch to obtain modulation weight. Correspondingly, the methods based on individual sample information can not maintain the contribution balance between bias-aligned and bias-conflicting samples, which is crucial for this problem as presented in Section~\ref{sec:quan_com}.
Different from the static rebalance method Rew, we propose a dynamic rebalance training strategy with aligned gradient contributions throughout the learning process, which enforces models to dive into intrinsic features instead of spurious correlations. Learning with GA, as demonstrated in Figure~\ref{fig:acc_c_mnist} and Table~\ref{tab:last_comp}, produces improved results with no degradation. The impact of GA on the learning trajectory presented in Figure~\ref{fig:grad_c_mnist} also shows that GA can schedule the optimization processes and take full advantage of the potential of different samples.
Besides, unlike the methods for class imbalance~\citep{cui2019class,tan2021equalization,zhao2020maintaining}, we try to rebalance the contributions of implicit groups rather than explicit categories.

\subsection{The sensitivity of the introduced hyper-parameters} 
\label{sec:hyperparameters}

Though the hyper-parameters are critical for methods aimed at combating unknown biases, recent studies~\citep{nam2020learning,kim2021learning} did not include an analysis for them. Here, we present the ablation studies on C-MNIST ($\rho=98\%$) for the hyper-parameters ($\eta$, $\tau$, $\gamma$) in our method as shown in Figure~\ref{fig:vary_param}. We find that the method performs well under a wide range of hyper-parameters. Specifically, for the confidence threshold $\eta$ in ECS, when $\eta$ $\rightarrow$ 0, most samples will be used to train the auxiliary biased models, including the bias-conflicting ones, resulting in low b-c scores for bias-conflicting samples (\textit{i.e.}, low recall of the mined bias-conflicting samples); when $\eta$ $\rightarrow$ 1, most samples will be discarded, including the relative hard but bias-aligned ones, leading to high b-c scores for bias-aligned samples (\textit{i.e.}, low precision). The determination of $\eta$ is related to the number of categories and the difficulty of tasks, \textit{e.g.}, 0.5 for C-MNIST, 0.1 for C-CIFAR10$^1$ and C-CIFAR10$^2$ (10-class classification tasks), 0.9 for B-Birds and B-CelebA (2-class) here. As depicted in Figure~\ref{fig:vary_param}, ECS achieves consistent strong mining performance around the empirical value of $\eta$. We also investigate ECS+GA with varying $\tau$. High precision of the mined bias-conflicting samples guarantees that GA can work in stage \uppercase\expandafter{\romannumeral2}, and high recall further increases the diversity of the emphasized samples. Thus, to ensure the precision first, $\tau$ is typically set to 0.8 for all experiments. From Figure~\ref{fig:vary_param}, ECS+GA is insensitive to $\tau$ around the empirical value, however, a too high or too low value can cause low recall or low precision, resulting in inferior performance finally. For the balance ratio $\gamma$, though the results are reported with $\gamma$ = 1.6 for all settings on C-MNIST, C-CIFAR10$^1$ and C-CIFAR10$^2$, 1.0 for B-Birds and B-CelebA, the proposed method is not sensitive to $\gamma$ $\in$ $[1.0, 2.0]$, which is reasonable as $\gamma$ in such region makes the contributions from bias-conflicting samples close to that from bias-aligned samples.

\begin{table}[t]
\centering
\caption{Average precision of the mined bias-conflicting samples on Colored MNIST ($\rho=98\%$).}
\begin{tabular}{cccc}
\toprule
\#auxiliary biased models & 1 & 2 & 4 \\
\midrule
AP (\%) & 95.0 & 98.8 & 99.0\\
\bottomrule
\end{tabular}%
\label{tab:num_biased_models}%
\end{table}%

We further add an ablation on the number of auxiliary models in Table~\ref{tab:num_biased_models}, showing more auxiliary biased models ($>2$) can get slightly better results. However, more auxiliary models will raise costs simultaneously, so we choose to use two auxiliary models in our design.

\begin{table}[t]
\centering
\caption{Accuracies (\%) on four test groups of Multi-Color MNIST. `$\infty$' states the reported results from DebiAN. The first line of the header \textit{w.r.t.} left color bias, the second one \textit{w.r.t.} right color bias.}
\setlength\tabcolsep{2pt}
\resizebox{\columnwidth}{!}{
\begin{tabular}{lccccc} 
\toprule
 & aligned & aligned & conflicting & conflicting & \multirow{2}[1]{*}{Avg.} \\
 & aligned & conflicting & aligned & conflicting &  \\
\midrule
LfF$^{\infty}$ & 99.6  & 4.7   & \textbf{98.6}  & 5.1   & 52.0  \\
PGI$^{\infty}$  & 98.6  & 82.6  & 26.6  & 9.5   & 54.3  \\
EIIL$^{\infty}$   & \textbf{100.0}  & \textbf{97.2}  & 70.8  & 10.9  & 69.7  \\
DebiAN$^{\infty}$ & \textbf{100.0}  & 95.6  & 76.5  & 16.0  & 72.0  \\
\midrule
ECS+GA  & \textbf{100.0}  & 89.7  & 96.1  & \textbf{24.3}  & \textbf{77.5}  \\
\bottomrule
\end{tabular}%
}
\label{tab:multi_bias}%
\end{table}%

\begin{table}[t]
\centering
\caption{Unbiased Accuracy (\%) on Colored MNIST with few bias-conflicting samples.}
\begin{tabular}{cccc}
\toprule
$\rho$ & 99.7\% & 99.9\% & 99.95\% \\
\#bias-conflicting samples & 180   & 60    & 30 \\
\midrule
Vanilla & 32.8 & 18.3 & 14.1 \\
Rew   & 56.0 & 27.3 & 22.9 \\
GA    & \textbf{68.0} & \textbf{60.0} & \textbf{53.9} \\
\bottomrule
\end{tabular}
\label{tab:rho_to1}
\end{table}

\subsection{When there are multiple biases} Most debiasing studies~\citep{nam2020learning,kim2021learning} only discussed single bias. However, there may be multiple biases, which are more difficult to analyze. To study the multiple biases, we further adopt the Multi-Color MNIST dataset following~\cite{Li_2022_ECCV} which holds two bias attributes: left color ($\rho=99\%$) and right color ($\rho=95\%$), see examples in Figure~\ref{fig:examples}. In such training sets, though it seems more intricate to group a sample as bias-aligned or bias-conflicting (as a sample can be aligned or conflicting \textit{w.r.t.} left color bias or right color bias separately), we still simply train debiased models with GA based on the b-c scores obtained via ECS. We evaluate ECS+GA on four test groups separately and present them in Table~\ref{tab:multi_bias}. We find the proposed method also can manage the multi-bias situation.

\begin{table}[t]
\centering
\caption{Accuracy (\%) on the unbiased training data (Colored MNIST, $\rho=10\%$).}
\begin{tabular}{ccc}
\toprule
Vanilla & LfF & ECS+GA  \\
\midrule
\textbf{97.8}  &   95.1  & 96.8 \\
\bottomrule
\end{tabular}
\label{tab:bal_train}
\end{table}

\begin{table}[t]
\centering
\caption{Performing GA with ordered learning on Colored MNIST ($\rho=98\%$). Here, ``Easy'' or ``Hard'' means only using bias-aligned or bias-conflicting training samples to update model.}
\resizebox{\columnwidth}{!}{
\begin{tabular}{cc|c}
\toprule
1-100  epochs  & 101-200  epochs & Unbiased  Accuracy (\%) \\
\midrule
Vanilla    &  Vanilla & 73.6 \\
Easy  & GA & 80.9  \\
Hard  & GA & 87.5 \\
GA    & GA & \textbf{89.1} \\
\bottomrule
\end{tabular}%
}
\label{tab:cu_learning}%
\end{table}%

\subsection{When there are only a few bias-conflicting samples}
\label{app:rho_analysis}

If the collected training set is completely biased (\textit{i.e.}, $\rho=100\%$), GA is not applicable. So, we want to know how GA performs when there are only a few bias-conflicting samples (\textit{i.e.}, $\rho \rightarrow 100\%$). The results are provided in Table~\ref{tab:rho_to1}, from which we find GA can achieve noticeable improvement even with few bias-conflicting samples.

\subsection{When training data is unbiased}
\label{app:safe}

It is important that the debiasing method is safe, \textit{i.e.}, can achieve comparable results to Vanilla when the training data is unbiased. We conduct experiments on Colored MNIST with $\rho=10\%$ (\textit{i.e.}, an unbiased training set) and the results are shown in Table~\ref{tab:bal_train}. From which, our method degrades slightly and still surpasses the debiasing method LfF. Actually, under an unbiased training set, our method tends to regard hard samples as bias-conflicting in stage \uppercase\expandafter{\romannumeral1}, and emphasize them in stage \uppercase\expandafter{\romannumeral2}.

\begin{figure*}[tb]
\centering
\includegraphics[width=0.93\textwidth]{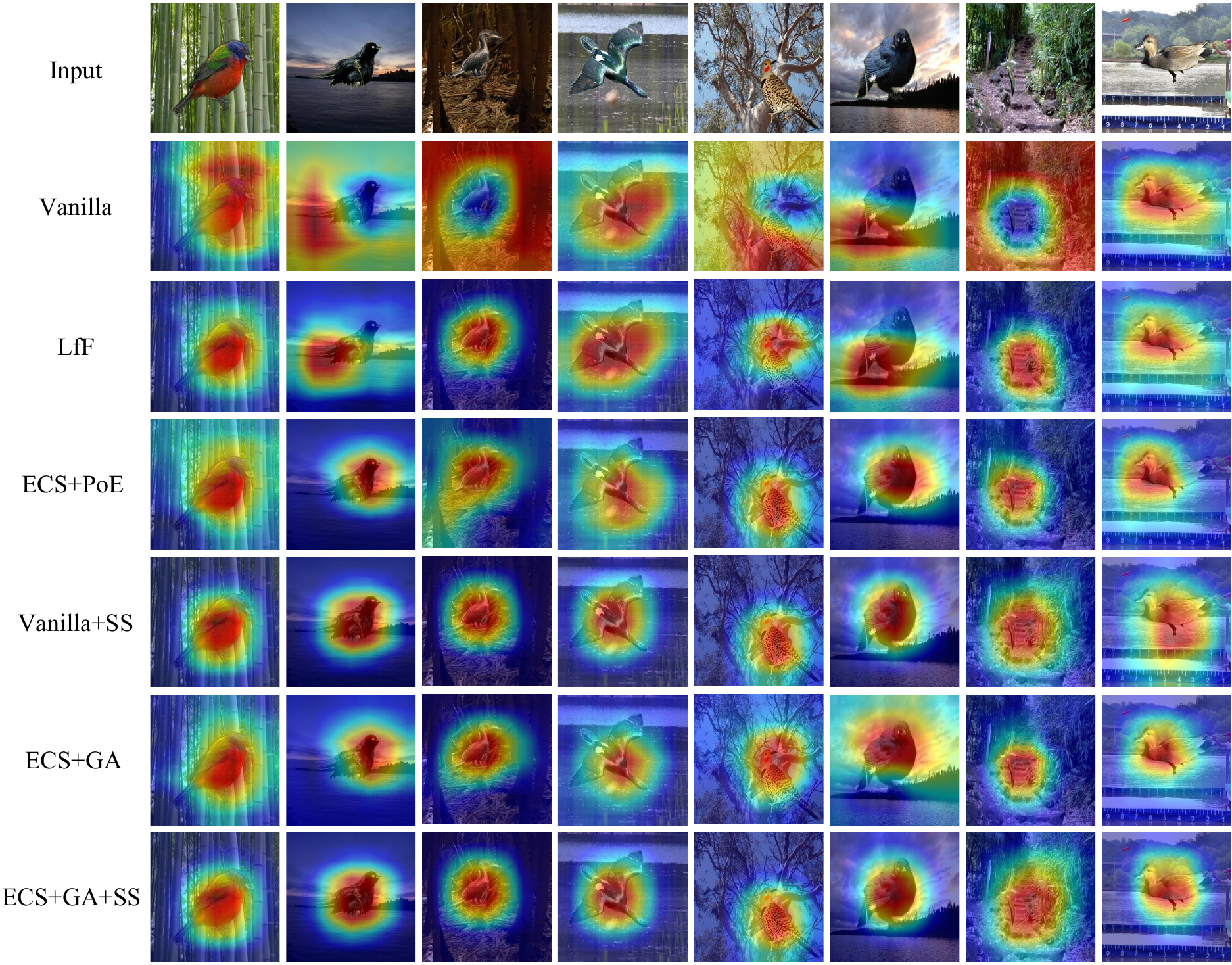}
\caption{Visualized activate maps of different models (last epoch) on Biased Waterbirds.}
\label{fig:cam_supp1}
\end{figure*}

\subsection{Connection to curriculum learning}
\label{app:curriculum}
Curriculum learning claims that using easy samples first can be superior, on the contrary, anti-curriculum learning argues that employing hard samples first is useful in some situations. We investigate the strategies of ordered learning in the context of debiasing. As presented in Table~\ref{tab:cu_learning}, both learning easy and hard samples first lead to inferior results than performing GA throughout the training process, showing that it is important to achieve a balance between bias-conflicting and bias-aligned samples in the whole learning stage.

\subsection{Visualization results}

We visualize the activation maps via CAM~\citep{zhou2016learning} in Figure~\ref{fig:cam_supp1}. Vanilla models usually activate regions related to biases when making predictions, \textit{e.g.}, the background in B-Birds. LfF and ECS+PoE can focus attention on key areas in some situations, but there are still some deviations. Meanwhile, the proposed ECS+GA and ECS+GA+SS mostly utilizes compact essential features to make decisions.

\section{Limitation and future work}
\label{sec:discussion}

Despite the achieved promising results, the debiasing method can be further improved in some aspects.
First, our method and many previous approaches (such as LfF, DFA, BiasCon, RNF-GT \textit{etc.}) are based on the assumption that there exist bias-conflicting samples in the training set. Although the assumption is in line with most actual situations, it should be noted that there are some cases where the collected training sets are completely biased (\textit{i.e.}, $\rho=100\%$), in which these methods are not applicable. For these cases, we should pay attention to methods that aim to directly prevent models from only pursuing easier features, such as SD.

Second, though the proposed ECS achieves significant improvement when compared with previous designs, we find that the bias-conflicting sample mining is not trivial, especially in complex datasets. The precision and recall achieved by our method on Biased Waterbirds and CelebA are still significantly lower than that on simple datasets like Colored MNIST and Corrupted CIFAR10 as shown in Table~\ref{tab:complete_pr}. For extreme cases, if the bias-conflicting scoring system fails, then the effect of GA can be influenced. Therefore, a better bias-conflicting scoring method is helpful and worth continuing to explore.

\section{Conclusions}
\label{sec:dis}
Biased models can cause poor out-of-distribution performance and even negative social impacts. In this paper, we focus on combating unknown biases which is urgently required in realistic applications, and propose an enhanced two-stage debiasing method. 
In the first stage, an effective bias-conflicting scoring approach containing peer-picking and epoch-ensemble is proposed. 
Then we derive a new learning objective with the idea of gradient alignment in the second stage, which dynamically balances the gradient contributions from the mined bias-conflicting and bias-aligned samples throughout the learning process. 
We also incorporate self-supervision into the second stage, assisting in the extraction of features. 
Extensive experiments on synthetic and real-world datasets reveal that the proposed solution outperforms previous methods.

\backmatter





\bmhead{Acknowledgments}

This work is supported in part by the National Natural Science Foundation of China under Grant 62171248, the R\&D Program of Shenzhen under Grant JCYJ20220818101012025, the PCNL KEY project (PCL2021A07), and Shenzhen Science and Technology Innovation Commission (Research Center for Computer Network (Shenzhen) Ministry of Education).

\newpage
\bibliography{sn-article}


\end{document}